\documentclass[10pt,letterpaper]{article}

\usepackage{cogsci-new}
\cogscifinalcopy 

\arxivcopy 

\usepackage[
    backend=biber,
    style=apa,
    natbib=true,
    doi=false,
    isbn=false,
    url=false,
    hyperref=false,
]{biblatex}
\usepackage{pslatex}
\usepackage{float}

\usepackage{amsmath}
\usepackage{amssymb}
\usepackage{graphicx}
\usepackage{xcolor}
\usepackage{xurl}
\usepackage{bbm}
\usepackage{booktabs}
\usepackage{xspace}
\usepackage{pifont}
\usepackage{enumitem}
\usepackage[bottom]{footmisc}
\usepackage{tabularx}
\usepackage{subcaption}
\usepackage{stmaryrd}
\usepackage{listings}
\usepackage{multirow}
\usepackage{xltabular}
\usepackage{epsdice}

\usepackage{tikz}
\usetikzlibrary{decorations.pathreplacing,calc}

\ifarxiv
\usepackage{geometry}
\usepackage{fancyhdr}
\fi

\usepackage{pgfplots}
\pgfplotsset{compat=newest}

\usepackage{url}
\usepackage{hyperref}
\hypersetup{
    colorlinks,
    linkcolor={blue!50!black},
    citecolor={blue!50!black},
    urlcolor={blue!80!black}
}

\usepackage[capitalise,nameinlink]{cleveref}
\creflabelformat{equation}{#2\textup{#1}#3} 
\crefformat{appendix}{#2#1#3}
\crefformat{section}{\S#2#1#3}
\crefname{algorithm}{Alg.}{Algs.}

\usepackage{listings}
\usepackage{inconsolata}

\usepackage[most]{tcolorbox}
\tcbsetforeverylayer{
  enhanced,
  fonttitle=\bfseries,
  fontupper=\ttfamily,
  left=1mm,
  right=1mm,
  top=1mm,
  bottom=1mm,
  boxsep=2mm, 
  colback=white,
  detach title,before upper={\tcbtitle\quad}
}

\definecolor{CustomLightBlue}{RGB}{173,216,230}
\definecolor{CustomLightGreen}{RGB}{138,234,146}

\newtcolorbox{systembox}{
  colback=gray!20!white,
  colframe=gray!80!black,
  coltitle=gray!80!black,
  title=System,
}
\newtcolorbox{userbox}[1][User]{
  colback=CustomLightBlue,
  colframe=CustomLightBlue!30!black,
  coltitle=CustomLightBlue!30!black,
  title=#1,
}
\newtcolorbox{assistantbox}{
  colback=CustomLightGreen!20!white,
  colframe=CustomLightGreen!50!black,
  coltitle=CustomLightGreen!50!black,
  title=Assistant,
}

\usepackage{etoolbox}
\let\clinenoskip\cline
\renewcommand{\cline}[1]{%
  \clinenoskip{#1}%
  \noalign{\smallskip}%
}

\usepackage{ragged2e}
\newcolumntype{Y}{>{\RaggedRight\arraybackslash}X}
\usepackage{array}

\newcommand{\prob}{p}
\newcommand{\entropy}{H}
\newcommand{\EIG}{\text{EIG}}

\newcommand{\board}{s}
\newcommand{\boards}{S}
\newcommand{\question}{x}

\newcommand{\lquestion}{l}
\newcommand{\answer}{y}
\newcommand{\answers}{Y_{\question}}

\renewcommand{\subsubsection}[1]{\textbf{#1} }

\title{Loose LIPS Sink Ships: \\ Asking Questions in \textit{Battleship} with Language-Informed Program Sampling}

\author{\textbf{Gabriel Grand}\normalfont{\textsuperscript{1,2}}\quad
\textbf{Valerio Pepe}\textsuperscript{2,3}\quad 
\textbf{Jacob Andreas}\textsuperscript{1}\quad
\textbf{Joshua B. Tenenbaum}\textsuperscript{1,2}\quad
\thanks{Correspondence to \texttt{gg@mit.edu}. Code for this paper is available at: \url{github.com/gabegrand/battleship}.} \\
\textsuperscript{1}MIT CSAIL \quad \textsuperscript{2}MIT BCS \quad \textsuperscript{3}Harvard SEAS\\
}

\begin{document}

\ifarxiv
\pagestyle{fancy}
\fancyhead{}
\fancyfoot{}
\fancyhead[L]{Published as a conference paper at \textit{CogSci 2024}.}
\fi

\maketitle

\begin{abstract}
\vspace{-0.5em}
Questions combine our mastery of language with our remarkable facility for reasoning about uncertainty. How do people navigate vast hypothesis spaces to pose informative questions given limited cognitive resources? We study these tradeoffs in a classic grounded question-asking task based on the board game \textit{Battleship}. Our language-informed program sampling (LIPS) model uses large language models (LLMs) to generate natural language questions, translate them into symbolic programs, and evaluate their expected information gain. We find that with a surprisingly modest resource budget, this simple Monte Carlo optimization strategy yields informative questions that mirror human performance across varied \textit{Battleship} board scenarios. In contrast, LLM-only baselines struggle to ground questions in the board state; notably, GPT-4V provides no improvement over non-visual baselines. Our results illustrate how Bayesian models of question-asking can leverage the statistics of language to capture human priors, while highlighting some shortcomings of pure LLMs as grounded reasoners.

\end{abstract}

\section{Introduction}

Human beings are question-generating machines.  From early childhood, we are driven to ask what, where, when, how and why.  But out of all the (infinitely many) grammatically-valid questions we could pose in a given situation, how do we decide which ones to ask?  And how do we find good questions efficiently, given such a large search space?

Questions can serve many functions, but a core goal is to gain information: reducing the speaker's uncertainty about the state of the world \citep{graesser1993exploring, markant2012does, hawkins2015you}. Informational value must be \textit{grounded} in a shared speaker-listener environment and is highly context-dependent.
For instance, ``Are you the guy in the red hat?'' is a natural question for Alice to text Bob in a crowded airport---but less so in a face-to-face interaction, or on a day when the local firefighter convention is in town.
Asking \textit{informative} questions therefore requires integrating linguistic competence with the ability to represent and reason about possible worlds.

In addition to grounding and context, question-asking is shaped by cognitive resource constraints \citep{anderson1990adaptive, chater1999ten, lieder2019resource}.
For instance, we know that both children and adults are ``greedy'' information-seekers in active learning and may consider only very few hypotheses at a time \citep{klayman1989hypothesis, vul2014one, markant2016self, meder2019stepwise, ruggeri2016sources}.
Faced with a cognitively demanding search task, people also prefer queries that yield simple answers that are easy to interpret \citep{cheyette2023people}.

In this paper, our goal is to model how people efficiently generate informative questions in a grounded environment, subject to resource constraints.  We explore several models in the context of the \textit{Battleship Game} \citep{rothe2017question, rothe2018people, rothe2019asking}: an adaptation of the classic board game to an open-ended question-asking task. Rothe et al. cast question-asking as program synthesis, where questions are expressed as symbolic programs in a domain-specific language (DSL).
They show that sampling programs and scoring them according to features learned from many human questions can approximate the distribution of questions people ask.

We aim to extend this approach in several ways.
Humans express questions via language---not code---and we would like models capable of the same. 
Nevertheless, symbolic programs are a useful format for expressing and evaluating questions; here, we bridge this gap by modeling meaning-making as the process of \textit{translating} from natural language to a language of thought (LoT).
Finally, questions can be cued by the situation; while previous models used the board strictly for top-down utility computation, here, the board also informs question generation  in a bottom-up way.

\begin{figure*}[ht!]
    \centering
    \includegraphics[width=0.9\textwidth]{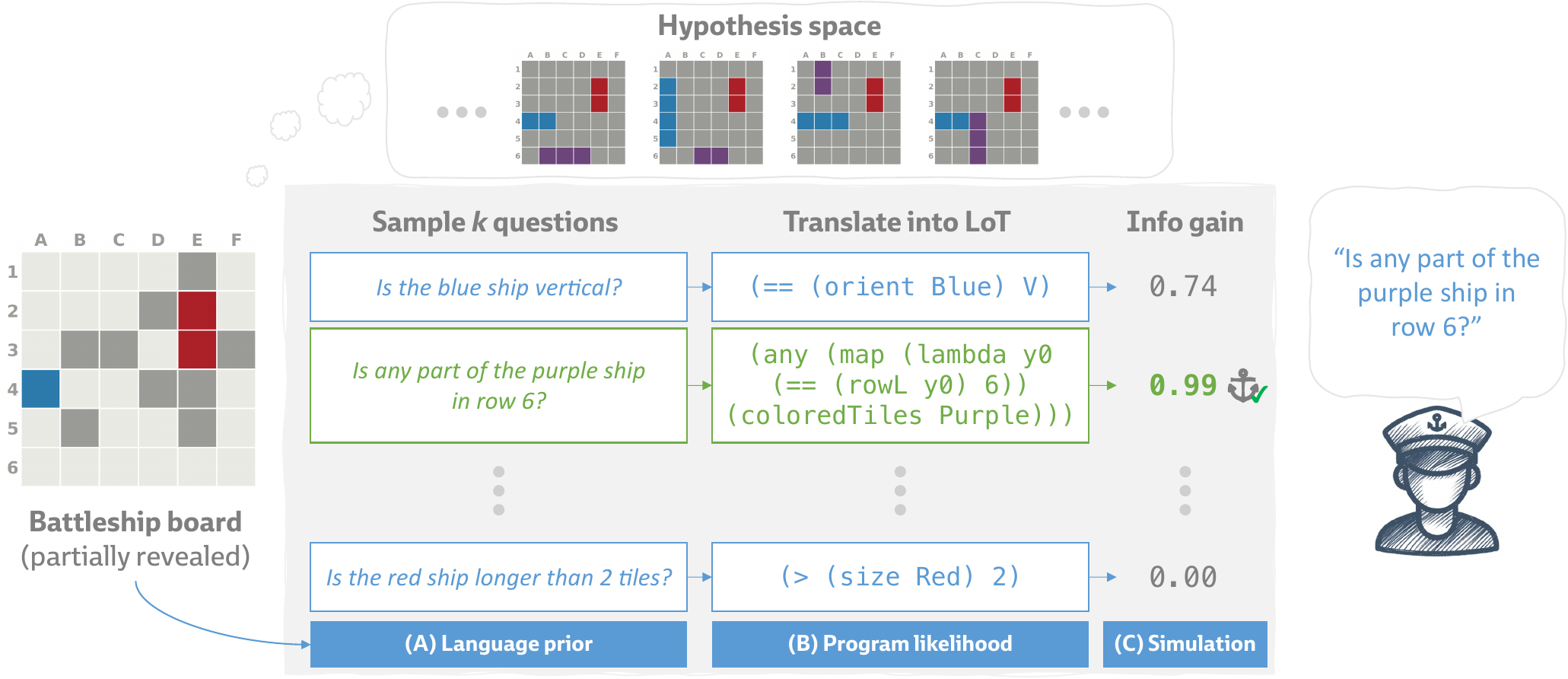}
    \caption{How do people formulate information-seeking questions in a grounded task such as the game \textit{Battleship}? Given a partially-revealed board, our LIPS model (A) samples $k$ questions from a language model prior and (B) translates these into LoT programs. (C) The utility of a question is computed by simulating the program against a hypothesis space of boards compatible with the observation. Here, the best question achieves Expected Information Gain (EIG) of 0.99, meaning the answer would rule out nearly half the boards in the hypothesis space. Our model is well-suited to filtering out samples from a noisy prior that are redundant (e.g., ``Is the red ship longer than 2 tiles?'') or inconsistent due to lack of grounding.
    }
    \label{fig:splash}
\end{figure*}

We build on a new approach for modeling language-informed thinking (Wong \& Grand et al., 2023) that integrates two powerful computational tools: large language models (LLMs) and probabilistic programs. LLMs allow our models to pose questions in free-form everyday language and translate those questions into symbolic representations. Probabilistic programs formalize the question-asker's world models and support coherent reasoning about the expected informativity of questions. Our work thus contributes to both Bayesian models of cognition and LLM accounts: We highlight how traditional models of active learning over structured hypothesis spaces can be extended to natural language settings, and how LLMs---which are increasingly being tuned to answer queries (e.g., \citealp{ouyang2022training})---can also be used to \textit{pose} questions that are coherent and grounded.

Our model (\cref{fig:splash}) is formulated as a simple Monte Carlo search that samples $k$ candidate questions stochastically from a prior distribution and estimates their informativity via simulation in an internal world model. LLMs play two distinct roles: (1) as a prior over questions, and (2) as a conditional distribution that maps questions from language into LoT programs. The translation step allows us to symbolically compute the Expected Information Gain (EIG) of candidate questions and choose the highest-value one. By varying $k$, we can control how much mental computation the model performs before producing a question. We call this overall framework ``Langugage-Informed Program Sampling'' (LIPS).

In our experiments, we compare question priors based on two different LLMs (CodeLlama-7b and GPT-4) as well as a probabilistic context-free grammar (PCFG) hand-engineered for the \textit{Battleship} domain. We find LLM priors yield informative questions that are well-calibrated to human data for surprisingly small values of $k$. In comparison, the PCFG requires slightly more samples to match mean human performance and yields a higher proportion of unnaturally complex questions. We also explore using LLMs as perceptual pattern learners to propose questions in a bottom-up manner. While a textual encoding of the world state does offer a moderate improvement in efficiency of question-asking, we find that a state-of-the-art multimodal LLM, GPT-4V, provides no improvement over non-visual baselines. Thus, the LLM-based models are also far from complete: they still struggle with grounding, producing many redundant or uninformative questions. In short, our results illustrate how cognitive models of informative question-asking can leverage LLMs to capture human-like priors, while highlighting some of the shortcomings of these models as grounded reasoners.

\section{The Battleship Game}

We adopt the \textit{Battleship} task developed by \citet{rothe2017question, rothe2018people}, a grid-based environment that evaluates participants' ability to ask goal-directed questions. In this task, participants are presented a partially-revealed board (\cref{fig:splash}) and asked to come up with a question that would help to reveal the location of the hidden ships. The task consists of 18 unique 6x6 board contexts, each containing three ships (red, blue, and purple) of varying length (2-4 tiles), orientation (horizontal or vertical), and placement. While later variants extended the paradigm to study multi-turn interactions \citep{rothe2019asking}, here we consider the original, single-turn task.

\section{Models}

Following prior work, we begin by considering an ideal observer model of a player that starts with a uniform prior $\prob(\board)$ over possible boards consistent with the observed initial state. After asking a question $\question$ and receiving an answer $\answer$, the player performs a Bayesian update to their belief distribution
\begin{equation}
\prob(\board \mid \answer; \question) = \frac{\prob(\answer \mid \board; \question) \prob(\board)}{\sum_{\board' \in \boards}{\prob(\answer \mid \board'; \question) \prob(\board')}}
\end{equation}
where the likelihood $\prob(\answer \mid \board; \question)$ 
is 1 if $\answer$ is consistent with $s$ and 0 otherwise. The marginal likelihood can be computed by enumeration or approximated by sampling over a hypothesis space of boards $\boards$.

The player's uncertainty about the hidden state of the game board can be measured by the Shannon entropy $\entropy(\board)$ \citep{shannon1948mathematical}, and the value of a question $\question$ can be defined as its Expected Information Gain (EIG):
\begin{equation}
\EIG(\question) = \entropy(\board) - \sum_{\answer \in \answers} \prob(\answer \mid \question) \entropy(\board \mid \question, \answer)
\label{eq:eig}
\end{equation}
Intuitively, EIG provides a log-space measure of the number of candidate boards that the player can rule out with question $\question$. For instance, an ideal yes/no question that rules out $50\%$ of possible boards would achieve $\EIG(\question)=1$. (Throughout, we use $\log_2$, so EIG is measured in bits.)

\begin{tikzpicture}
   \begin{axis}[
       xlabel={Remaining board hypotheses},
       ylabel={$\EIG(\question)$},
       xmin=0, xmax=10,
       ymin=0, ymax=4,
       axis lines=center,
       axis on top=true,
       domain=0.1:10,
       grid=major,
       grid style={dashed,gray!30},
       samples=100,
       smooth,
       width=\linewidth,
       height=0.45\linewidth,
       xtick={1,2,3,4,5,6,7,8,9,10},
       xticklabels={$100\%$,$\text{\quad}50\%$,,$\text{\quad}25\%$,,,,$\text{\quad}12.5\%$,,,}
   ]
   \addplot [blue,thick] {log2(x)};
   \end{axis}
\end{tikzpicture}
\noindent In \textit{Battleship}, questions that admit a large set of possible answers, denoted $\answers$, can achieve $\EIG(\question) \gg 1$ (e.g., ``What is the top-left corner of the red ship?''). However, some answers may be more informative than others; this uncertainty gives rise to the expectation over possible answers in \cref{eq:eig}.

In prior work, EIG was considered as one among several heuristic features (complexity, answer type, etc.) in a Boltzmann energy model that was fit to maximize the likeihood of the collected human questions \citep{rothe2017question}. Here we take a complementary approach: instead of fitting our model to human data collected from \textit{Battleship}, we instead aim to sample directly from a distribution of maximally-informative questions---\textit{without} positing the space of features these questions might have. We hypothesize that human-like questions will fall out naturally from a Bayesian model with a very generic prior that is subject to cognitive resource constraints. 

We formulate our model as a probabilistic sample-based search with a parameter $k$ that controls the amount of internal computation the model performs. (We are in part inspired by the bounded space model of \citealp{ullman2016coalescing} for creative language generation.) Given some proposal distribution over questions, we sample $k$ questions and choose the one that maximizes EIG:
\begin{align}
\{ \question_1, \ldots, \question_k \} &\sim \prob (\question \mid \board) \\
\question^* &= \arg\max_{\question_i} \EIG(\question_i)
\end{align}
A central challenge of this approach is choosing a suitable proposal distribution $\prob (\question \mid \board)$ that admits efficient sampling. Moreover, as the notation implies, this distribution should ideally be \textit{board-conditional} so as to generate targeted questions about the particular board at hand. To facilitate computation of EIG, it is also desirable to have a proposal distribution that is capable of expressing questions as LoT-like programs that can be deterministically executed against the board following some denotational semantics; i.e., $\answer = \llbracket \question \rrbracket_\board$. We consider two kinds of question-proposal distribution that allow us to instantiate our LIPS model.

\subsection{Grammar proposal distribution}

We begin by considering a probabilistic context-free grammar(\citealp{johnson1998pcfg}) as a proposal distribution over questions. We adopt the grammar of~\citet{rothe2017question}\footnote{See Table SI-1 in \citet{rothe2017question} for the full grammar. We omit $\lambda$-abstractions, which rarely yield well-formed questions during sampling, and we filter out trivial expressions of depth 1.} whose rules and terminals correspond to key concepts in \textit{Battleship}: ships vary in \textit{color}, \textit{size}, \textit{orientation}, \textit{location}, etc. The grammar also encodes numeric and set-theoretic operations to support comparisons; e.g., ``How many of the blue ship's tiles are in column B?''

\begin{small}
\begin{align*}
    \textbf{Answer} &\rightarrow \text{Bool} \mid \text{Num.} \mid \text{Color} \mid \text{Orient.} \mid \text{Loc.} \\
    \text{Bool} &\rightarrow \text{`T'} \mid \text{`F'} \mid \text{(and B B)} \mid \text{(touch Ship Ship)} \ldots \\
    \text{Num.} &\rightarrow 0 \mid 1 \mid \ldots \mid 9 \mid \text{(+ N N)} \mid \ldots \\ 
    \text{Num.} &\rightarrow \text{(size Ship)} \mid \text{(row L)} \mid \text{(col L)} \ldots \\
    \text{Color} &\rightarrow \text{Ship} \mid \text{} \\
    \text{Ship} &\rightarrow \text{`Blue'} \mid \text{`Red'} \mid \text{`Purple'} \\ 
    \text{Orient.} &\rightarrow \text{`Horizontal'} \mid \text{`Vertical'} \mid \text{(orient Ship)} \\
    \text{Loc.} &\rightarrow 1A \mid 1B \mid \ldots \mid 6F \mid \text{(topleft Set)} \ldots \\
    \text{Set} &\rightarrow \text{(tiles Color)} \mid \text{(}\cap\text{Set Set)} \mid \text{`AllColors'} \ldots
\end{align*}
\end{small}

As a computational instantiation of a cognitive theory, the PCFG proposal distribution follows in the (probabilistic) language of thought tradition \citep{fodor1975language, goodman2014concepts, goodman2015probabilistic}. One advantage of this formalization is that it comes with clear-cut denotational semantics: questions are programs that can be executed against the board state to yield an answer. Additionally, the PCFG imparts an inductive bias towards shorter programs that naturally implements Bayesian Occam's Razor \citep{gelman1995bayesian, henderson2010structure}. However, the PCFG also yields a combinatorially-large number of trivial statements (e.g., mathematical propositions that do not make reference to the board). These issues could be addressed by making the PCFG board-conditional---e.g., by training a neural  ``recognition network'' \citep{ellis2021dreamcoder} to map board inputs to weights. Nevertheless, achieving a good fit with this model would require collecting a labeled dataset containing many thousands of $(\board, \question)$ pairs. In the absence of such data, we follow the prior work in treating the PCFG as an unconditional prior with uniform probability over the production rules. 

\subsection{Language model proposal distribution}
A recent line of work in probabilistic programming explores using Large Language Models (LLMs) as instantiations of humanlike priors in Bayesian models \citep{lew2020leveraging, lew2023sequential, dohan2022language, ellis2023human}. For the purpose of constructing a cognitive model of human question-asking, LLMs represent an attractive proposal distribution for several reasons. First, since they are trained on vast corpora of natural text, LLMs directly encode a prior over plausible questions. Moreover, LLMs are strong in-context learners \citep{brown_language_2020} and are increasingly amenable to instruction from the experimenter \citep{ouyang2022training, rafailov2024direct}. Consequently, by constructing an appropriate prompt, we can transform a generic LLM into a proposal distribution over questions in the \textit{Battleship} domain. This approach faces two main challenges, which we detail below.

\subsubsection{Grounding generation in the state of the world}
\begin{figure}
    \centering
    \includegraphics[width=\linewidth]{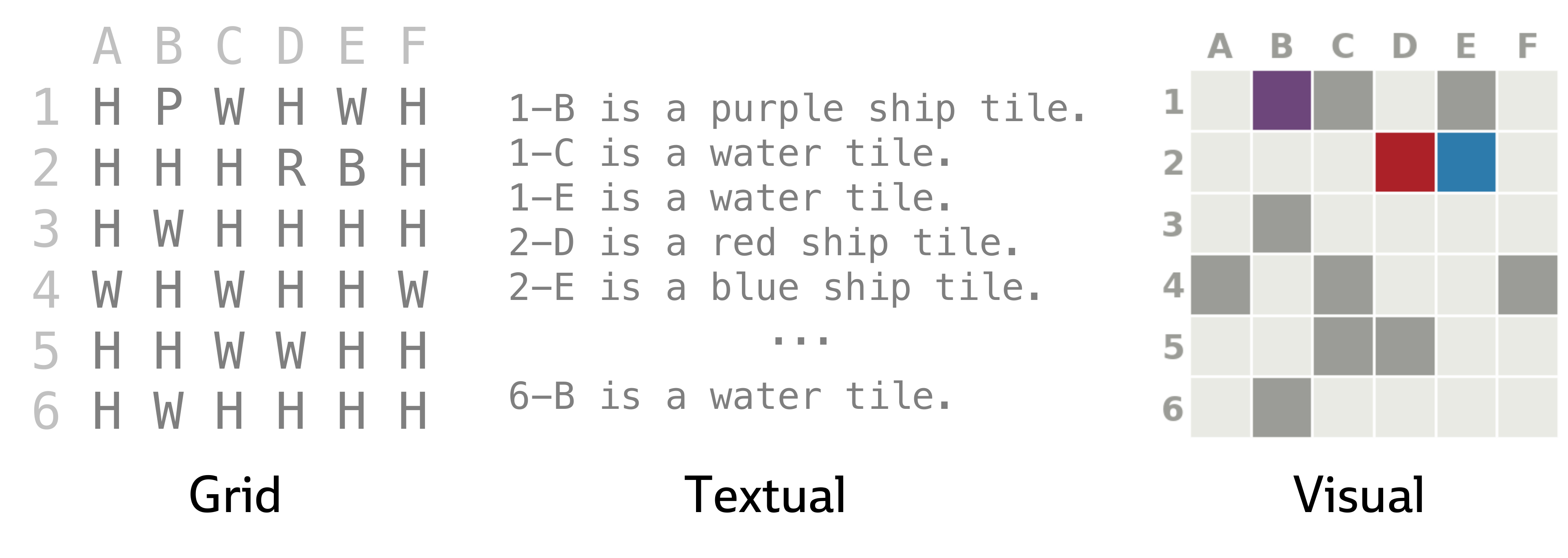}
    \caption{We experiment with 3 different board representations: an ASCII-style grid, a textual serialization, and a visual prompt encoded as an image.}
    \vspace{-1em}
    \label{fig:board_representation}
\end{figure}

Ideally, we would like our model to be ``stimulus computable'' \citep{yamins2016using}, accepting the same images and task instructions as a human participant. While multimodal LLMs are growing in popularity and availability \citep{driess2023palm, openai2023gpt4vision}, it remains unclear to what extent they are capable of extracting structured visual information---such as a \textit{Battleship} board---into an appropriate computational representation. We experiment with three different types of board representation (\cref{fig:board_representation}) in order to evaluate the degree to which our LLM proposal distributions are able to leverage board-conditional information.

\subsubsection{Translating from natural language to the LoT} Our LIPS model posits that the question-asker mentally draws and evaluates $k$ samples and chooses the most informative one. This is straightforward in the case of the PCFG, which directly generates programs, but not for the case of LLMs, which output natural language. To address this, we follow the approach of the \textit{Rational Meaning Construction} framework (Wong \& Grand et al., 2023), which uses LLMs to implement a ``meaning function'' that translates from natural language into the LoT. Concretely, we decompose the LLM proposal into separate \textbf{linguistic question generation} $\prob(\lquestion \mid \board)$ and \textbf{language-to-program translation} $\prob(\question \mid \lquestion)$ distributions, which we approximate via sampling.
\begin{equation}
\prob(\question \mid \board) = \sum_{\lquestion} \prob(\question \mid \lquestion) \prob(\lquestion \mid \board)
\label{eq:translation}
\end{equation}
This formalization admits many possible denotational semantics---$\llbracket \cdot \rrbracket_s$ could be implemented by a LISP interpreter, a Python program, or even a LLM. For convenience, we use the same \textit{Battleship} DSL from \citet{rothe2017question}, which allows us to take advantage of the fast C++ implementation of the EIG function developed for that work.\footnote{\url{https://github.com/anselmrothe/EIG}} 

\section{Experiment}

\subsection{Participants, materials, and methods}

\subsubsection{Human data}
We use the human dataset collected by \citealt{rothe2017question}, which consists of 26-39 questions for each board composed by a single pool of N=40 participants, for a total of 605 question-board pairs. Participants were not ``prompted'' with any example questions; they were only given the constraint that the question should admit a single-word answer. As the program annotations in this dataset used an earlier version of the DSL, we manually translated a representative subset of the questions into the latest DSL and used a LLM to annotate the remaining programs.

\subsubsection{LLMs}
We queried GPT-4 \citep{openai2023gpt4, openai2023gpt4vision} via API, using \texttt{gpt-4-0613} for the textual and grid board formats, and \texttt{gpt-4-vision-preview} for the visual format. To compare against a reproducible, open-source LLM, we used CodeLlama \citep{roziere2023code}, a member of the Llama~2 family of models that was finetuned for code generation. We obtained the model weights from HuggingFace (\texttt{CodeLlama-7b-hf}) and used the smallest variant of the model, which contains 7B parameters. We performed local inference on a single GPU, taking advantage of the \texttt{hfppl} library \citep{lew2023sequential} to speed up inference via caching.

\subsubsection{Prompting}
\ifarxiv
We fed both LLMs identical sets of algorithmically-constructed prompts (see the ``Prompts'' section in the Appendix).
\else
We fed both LLMs identical sets of algorithmically-constructed prompts.
\fi
For question generation $\prob(\lquestion \mid \board)$, each prompt consisted of instructions describing the task setup (``You are playing the board game Battleship. There are three ships on the board...''). In the \textbf{zero-shot} condition, the prompt concluded with a target game board (\cref{fig:board_representation}) and text to elicit a question. In the \textbf{few-shot} condition, the prompt additionally included 3 example boards, each with 10 questions from the human data. The example boards and questions were sampled without replacement in a leave-one-out manner so as to exclude human data collected for the target board. For translation $\prob(\question \mid \lquestion)$, the prompt consisted of a similar task instruction, followed by 12 $(\lquestion, \question)$ pairs randomly sampled from the human data in the same manner. 

\subsubsection{Sampling}
For each LLM condition, we sampled 100 questions/board $\times$ 18 boards. To explore the effects of prompt and board formats, we repeated this process for each combination of \{zero-shot, few-shot\} $\times$ \{textual, grid, visual, no board\} using GPT-4(V). For the PCFG, which is not board-conditional, we sampled a single set of 100K questions and computed their EIG values for each board. Following \citet{ullman2016coalescing}, to avoid expensive re-collection of data, samples were grouped \textit{post-hoc} into buckets of size $k$. Since the underlying samples are i.i.d., this provides an unbiased estimate of the true sampler, with the caveat that the effective sample size diminishes with $k$. Throughout, null hypothesis testing was conducted between conditions using Welch's t-test.

\begin{figure*}[htbp!]
    \centering
    \begin{subfigure}[b]{0.49\linewidth}
        \centering
        \includegraphics[width=\linewidth]{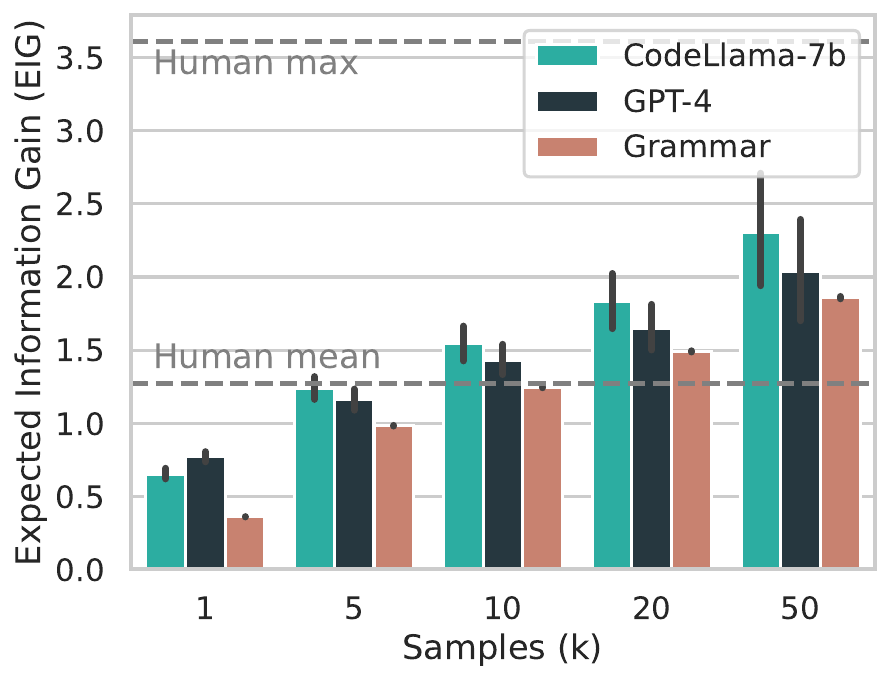}
    \end{subfigure}
    \begin{subfigure}[b]{0.49\linewidth}
        \centering
        \includegraphics[width=\linewidth]{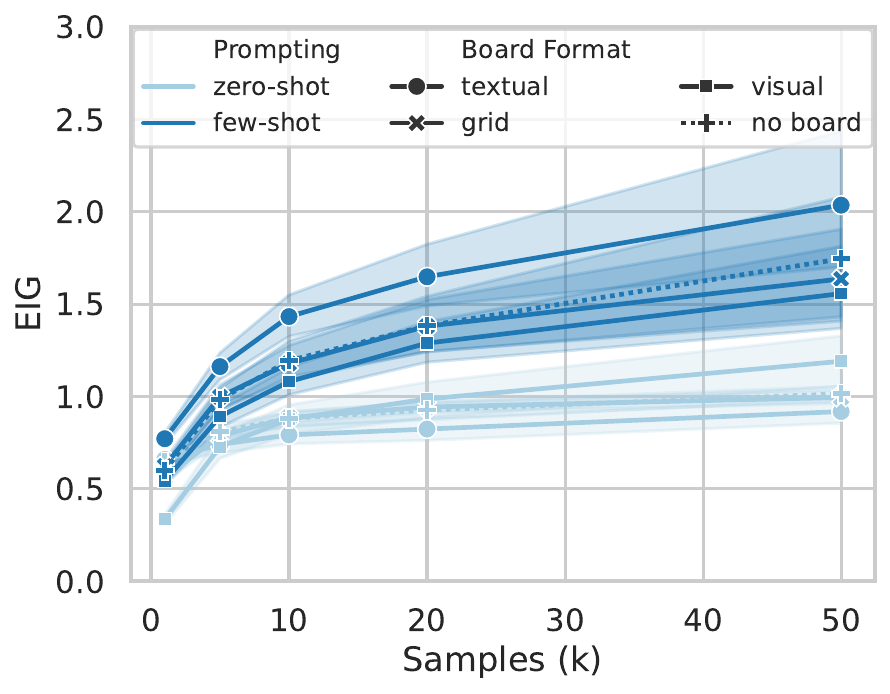}
    \end{subfigure}

    
    \begin{subfigure}[b]{\linewidth}
        \centering
        \includegraphics[width=\linewidth]{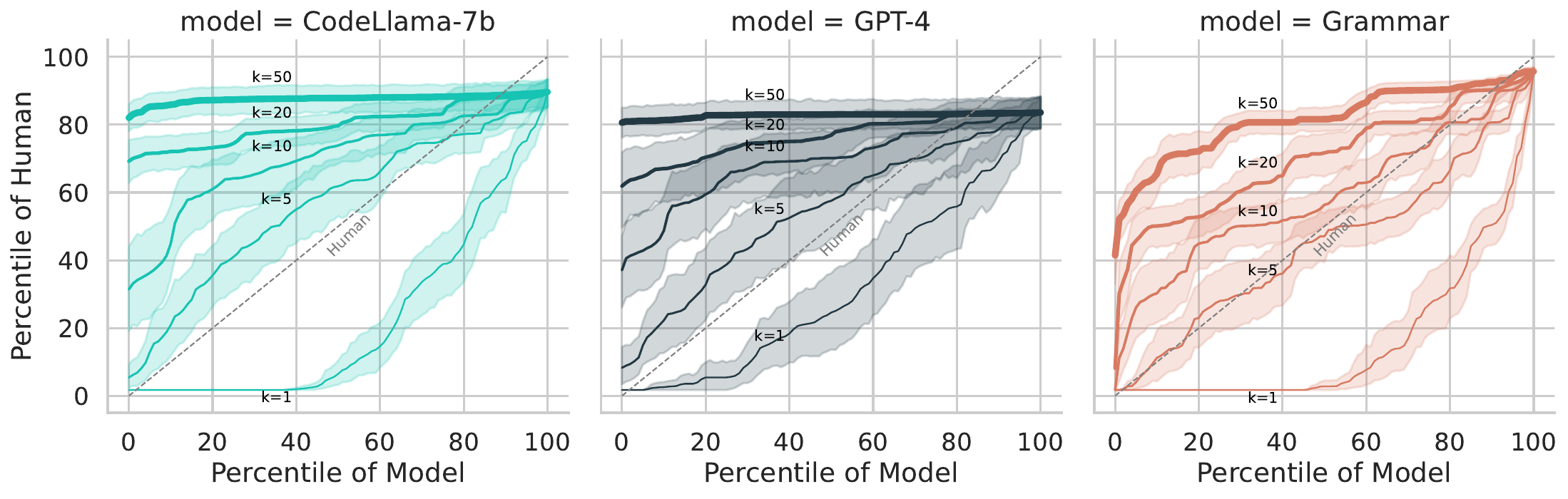}
    \end{subfigure}

    
    \caption{Comparing the informativity of model-generated questions against human data. \textbf{(Top left)} LIPS with two LLMs and a hand-engineered grammar as proposal distributions over questions. As $k$ increases, all three models reach mean-human performance, though they fall short of the best human-generated questions.
    \textbf{(Top right)} Evaluating GPT-4's performance with different prompt formats and board representations. Including few-shot examples universally boosts EIG. However, performance varies depending on the board format. Notably, GPT-4(V) was unable to utilize the board's structure in text (grid) or images (visual), implying a failure of grounding.
    \textbf{(Bottom)} Q-Q plots comparing model vs. human EIG values at varying sample sizes. At $k=5$, all three models are generally well-calibrated to humans, though they fall short of the top 10-20\% of human questions. Throughout, error bars and shaded regions indicate 95\% bootstrapped confidence intervals. GPT-4 and CodeLlama-7b refer to the few-shot, textual condition unless otherwise noted.
    }
    \label{fig:main_results}
\end{figure*}

\begin{table*}[htbp!]
\centering
\footnotesize
\begin{tabular}{lrrrrrrrrrrrr}
\toprule
Model & \multicolumn{2}{c}{EIG} & \multicolumn{2}{c}{\% Valid} & \multicolumn{2}{c}{\% Informative} & \multicolumn{2}{c}{Program Depth} & \multicolumn{2}{c}{Program Size} & \multicolumn{2}{c}{Question Words} \\
 & $\mu$ & $\sigma_M$ & $\mu$ & $\sigma_M$ & $\mu$ & $\sigma_M$ & $\mu$ & $\sigma_M$ & $\mu$ & $\sigma_M$ & $\mu$ & $\sigma_M$ \\
\midrule
Human & 1.27 & 0.04 & 1.00 & 0.00 & 0.97 & 0.01 & 3.22 & 0.07 & 4.51 & 0.14 & 7.12 & 0.08 \\
Grammar & 0.36 & 0.00 & 1.00 & 0.00 & 0.38 & 0.00 & 3.01 & 0.00 & 5.13 & 0.01 & -- & -- \\
CodeLlama-7b & 0.65 & 0.02 & 0.75 & 0.01 & 0.45 & 0.01 & 2.64 & 0.02 & 3.24 & 0.04 & 6.66 & 0.04 \\
GPT-4 (few-shot) & 0.77 & 0.02 & 0.88 & 0.01 & 0.59 & 0.01 & 2.61 & 0.02 & 3.22 & 0.04 & 6.23 & 0.03 \\
GPT-4 (zero-shot) & 0.66 & 0.01 & 0.40 & 0.01 & 0.35 & 0.01 & 3.73 & 0.04 & 5.04 & 0.09 & 5.19 & 0.02 \\
GPT-4 (no board) & 0.60 & 0.02 & 0.68 & 0.01 & 0.43 & 0.01 & 3.08 & 0.03 & 4.12 & 0.07 & 6.28 & 0.03 \\
\bottomrule
\end{tabular}

\caption{Summary statistics of the underlying samples ($k=1$) across all board contexts. Questions that translated to a parseable program are considered Valid, and those that achieved $\EIG > 0$ are considered Informative. Program Depth and Size refer to the depth and number of nodes of the program abstract syntax tree. Question Words measures the number of words in the natural language question. $\mu$ and $\sigma_M$ denote sample mean and standard error, respectively.}
\label{tab:main_results}
\end{table*}

\subsection{Results and Discussion}

\subsubsection{Informativity} How informative are the questions collected from humans? And to what extent do our models capture the information-seeking quality of human questions? We computed EIG values for all human and model-generated questions (\cref{tab:main_results}). Across the 18 boards, the average human question scored $\EIG=1.27$, while the best human question achieved considerably higher $\EIG=3.61$. Despite this large range, \textit{virtually all} ($97\%$) of the human questions were informative ($\EIG>0$), revealing that participants were highly sensitive to the board state.

In contrast, the underlying proposal distributions $(k = 1)$ were substantially noisier than people: questions from CodeLlama and GPT-4 averaged $\EIG=0.65$-$0.66$, respectively, while questions from the grammar averaged $\EIG=0.36$. However, as \cref{fig:main_results} (top left) reveals, LIPS allows for a significant boost in performance: with just $k=5$ samples, both LLMs approached human mean performance; and at $k=10$, both models significantly outperformed the human mean, with $p<0.001$ for CodeLlama, and $p=0.01$ for GPT-4 (textual, few-shot). This trend continues for sample sizes $k=20$ and $k=50$, though all models still fall short of the best human-generated questions.

\subsubsection{Sample efficiency} What represents a cognitively-plausible amount of mental sampling? \cref{fig:main_results} (bottom) compares the full distribution of model vs. human EIG values for varying values of $k$. At $k=5$, both LLMs were closely calibrated to the human distribution, performing on par with the grammar, which was hand-engineered to capture this distribution. In other words, the $N$th percentile of human question-askers wrote questions that were of comparable informativity to the $N$th percentile of samples from the model. However, the top human questions (approx. 85-90th percentile) outperformed the top model-generated questions.

\subsubsection{Translation fidelity} One restriction of our evaluation is that, in order for a question to be considered informative, it needs to be expressable in the \textit{Battleship} DSL. But how effective is the model at translating questions into programs? As \cref{tab:main_results} (\% Valid) shows, a high percentage of samples from CodeLlama (75\%) and GPT-4 (88\%) were successfully translated. Only in the GPT-4 (zero-shot) case did the translation model achieve low fidelity (40\%). Since the model does not receive any examples in the zero-shot case, it is not surprising that many of the questions from this distribution were not translatable.

\subsubsection{Groundedness} To what extent did the LLM-generated questions take the board state into account? Of the valid programs sampled from each model, 40\% (CodeLlama) and 33\% (GPT-4) were \textit{uninformative} ($\EIG=0$). This occurs when a question is redundant with respect to information already revealed in the board. (For instance, ``Is the red ship vertical?'' is uninformative for 3/18 boards in the stimulus set.) The high proportion of uninformative programs highlights a potential failure of grounding. Our evaluation of different board formats, shown in \cref{fig:main_results} (top right), provides further evidence of this issue. Of the four board formats (\cref{fig:board_representation}),  the ``textual'' representation was the only one that significantly outperformed the ``no board'' condition ($p<0.05$ for $k=1$-$20$). Notably, across $k$, the ``visual'' board format performed either significantly worse ($p<0.05$ for $k=1,5$) or was not significantly different than the ``no board'' condition ($p>0.05$ for $k=10$-$50$). These results show that that GPT-4V was unable to utilize the board's structure to formulate informative questions relative to a board-agnostic baseline.

\subsubsection{Question type} 
\begin{figure}
    \centering
    \includegraphics[width=\linewidth]{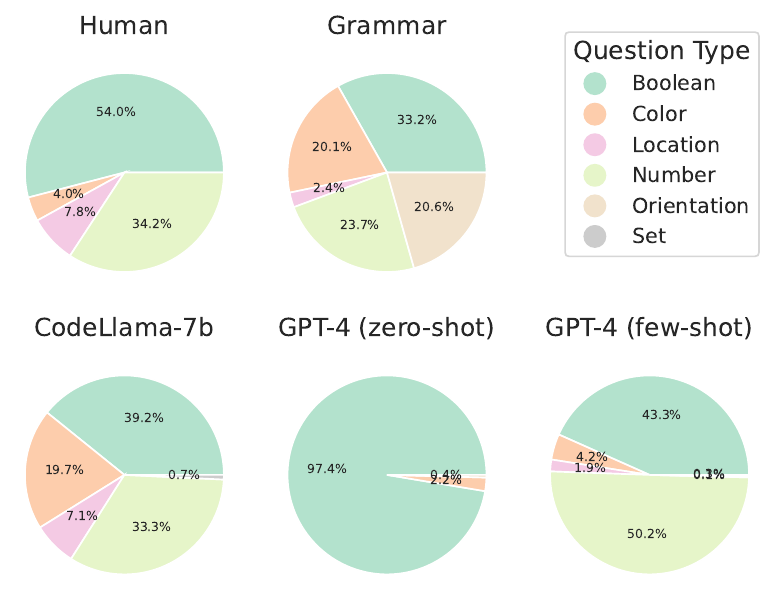}
    \caption{Proportion of top-level question types generated by each proposal distribution at $k=1$.}
    \label{fig:question_types}
    \vspace{-1em}
\end{figure}

What \textit{kinds} of information do humans ask about, and do the models reflect this distribution? As illustrated in \cref{fig:question_types}, humans ask a diverse range of question types, with a preference for boolean and numeric answers. Owing to its structure, the grammar generates an approximately uniform distribution over types. Meanwhile, both of the few-shot prompted LLMs approximate the human distribution, though CodeLlama mirrors it more closely than GPT-4. Without access to examples, GPT-4 (zero-shot) defaults to boolean questions that echo traditional \textit{Battleship} moves; e.g., ``Is there a ship at 2-C?'' Thus, the different choices of prior encode different inductive biases---and LLMs provide an especially flexible way of encoding both human general knowledge and domain-specific priors into cognitive models.

\section{Conclusion}

As more and more people interact with language models on a daily basis, understanding how \textit{humans} seek information through language is a truly important scientific question. In this work, we introduced a new approach to modeling informative question-asking by sampling questions from a noisy LLM prior and translating into programs in a LoT. But where does this LoT come from in the first place? Here, we used an existing DSL as initial step, but our approach could be combined with Bayesian program induction techniques to learn a new DSL from data \citep{ellis2021dreamcoder, wong2021leveraging, grand2024lilo, piantadosi2024learning}. Relaxing our assumptions even further, we might eschew a DSL in favor of a domain-general programming language like Python \citep{ellis2023human, wang2024hypothesis}.

While Monte Carlo sampling is attractive for its simplicity, there exist more sophisticated inference techniques that offer better sample efficiency. Several such approaches have recently been studied in an AI dialogue context for eliciting user preferences \citep{piriyakulkij2023active, li2023eliciting} and clarifying ambiguity \citep{zhang2023clarify}. Applying these inference methods to study people's behavior across longer interactions---such as multi-turn \textit{Battleship}---presents a natural direction for future work.

\clearpage
\newpage
\section{Acknowledgments}

We thank Brenden Lake, Maddy Bowers, Lionel Wong, Sam Cheyette, and Guy Davidson for helpful discussions and feedback on this work. 

The authors gratefully acknowledge support from the MIT Quest for Intelligence, the MIT-IBM Watson AI Lab, the Intel Corporation, AFOSR, DARPA, and ONR. GG is supported by the National Science Foundation (NSF) under Grant No. 2141064. JA is supported by NSF Grant IIS-2144855. JBT received support from AFOSR Grant \#FA9550-19-1-0269, the MIT-IBM Watson AI Lab, ONR Science of AI and the DARPA Machine Common Sense program. Any opinions, findings, and conclusions or recommendations expressed in this material are those of the author(s) and do not necessarily reflect the views of sponsors.

\nocite{wong2023word}
\printbibliography 

\ifarxiv

\clearpage
\newpage

\onecolumn
\appendix
\section{Appendix}

\begin{xltabular}{\textwidth}{cllXYr}
\caption{Examples questions and programs for each board context. For each model, we sampled one best $\star$ and one random $\epsdice{6}$ question. For humans, the random sample was drawn from the full pool of participant data for each trial; for models, the random sample was selected from the LIPS outputs with $k=10$, which provides a close match to mean human performance. In cases where the best EIG value was attained by multiple programs, tiebreaking was random. Note that the Grammar generates programs directly---many of which are not readily translatable to natural language---so ``Question'' is omitted for this model.} \label{apx:tab:examples} \\
\toprule
Board & Model & & Question & Program & EIG \\
\midrule
\endfirsthead
\toprule
Board & Model & & Question & Program & EIG \\
\midrule
\endhead
\endfoot
\bottomrule
\endlastfoot
\multirow[]{8}{*}{\shortstack[m]{Trial 1 \\ \includegraphics[width=1in]{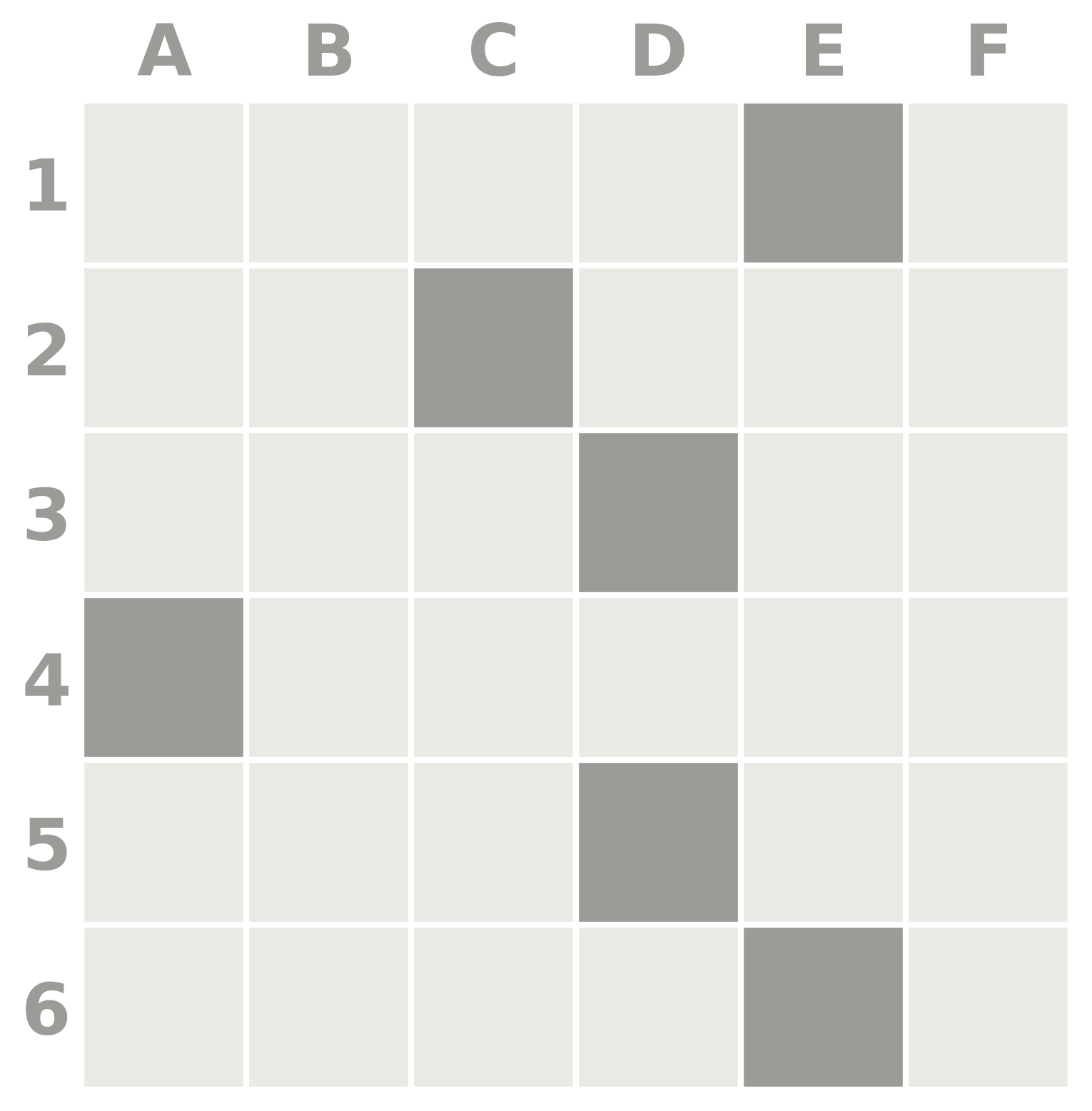}}} & \multirow[]{2}{*}{Human} & $\star$ & At what location is the top left part of the red ship? & \texttt{(topleft (coloredTiles Red))} & 4.67 \\
 &  & $\epsdice{6}$ & Is there a purple tile at 1A? & \texttt{(== (color 1A) Purple)} & 0.39 \\
\cline{2-6}
 & \multirow[]{2}{*}{CodeLlama} & $\star$ & At what location is the top left part of the blue ship? & \texttt{(topleft (coloredTiles Blue))} & 4.67 \\
 &  & $\epsdice{6}$ & How many tiles is the blue ship? & \texttt{(size Blue)} & 1.36 \\
\cline{2-6}
 & \multirow[]{2}{*}{GPT-4} & $\star$ & What is the location of one blue tile? & \texttt{(topleft (coloredTiles Blue))} & 4.67 \\
 &  & $\epsdice{6}$ & How many tiles is the blue ship? & \texttt{(size Blue)} & 1.36 \\
\cline{2-6}
 & \multirow[]{2}{*}{Grammar} & $\star$ & --- & \texttt{(topleft (coloredTiles Red))} & 4.67 \\
 &  & $\epsdice{6}$ & --- & \texttt{(color 6B)} & 1.40 \\
\cline{1-6}
\multirow[]{8}{*}{\shortstack[m]{Trial 2 \\ \includegraphics[width=1in]{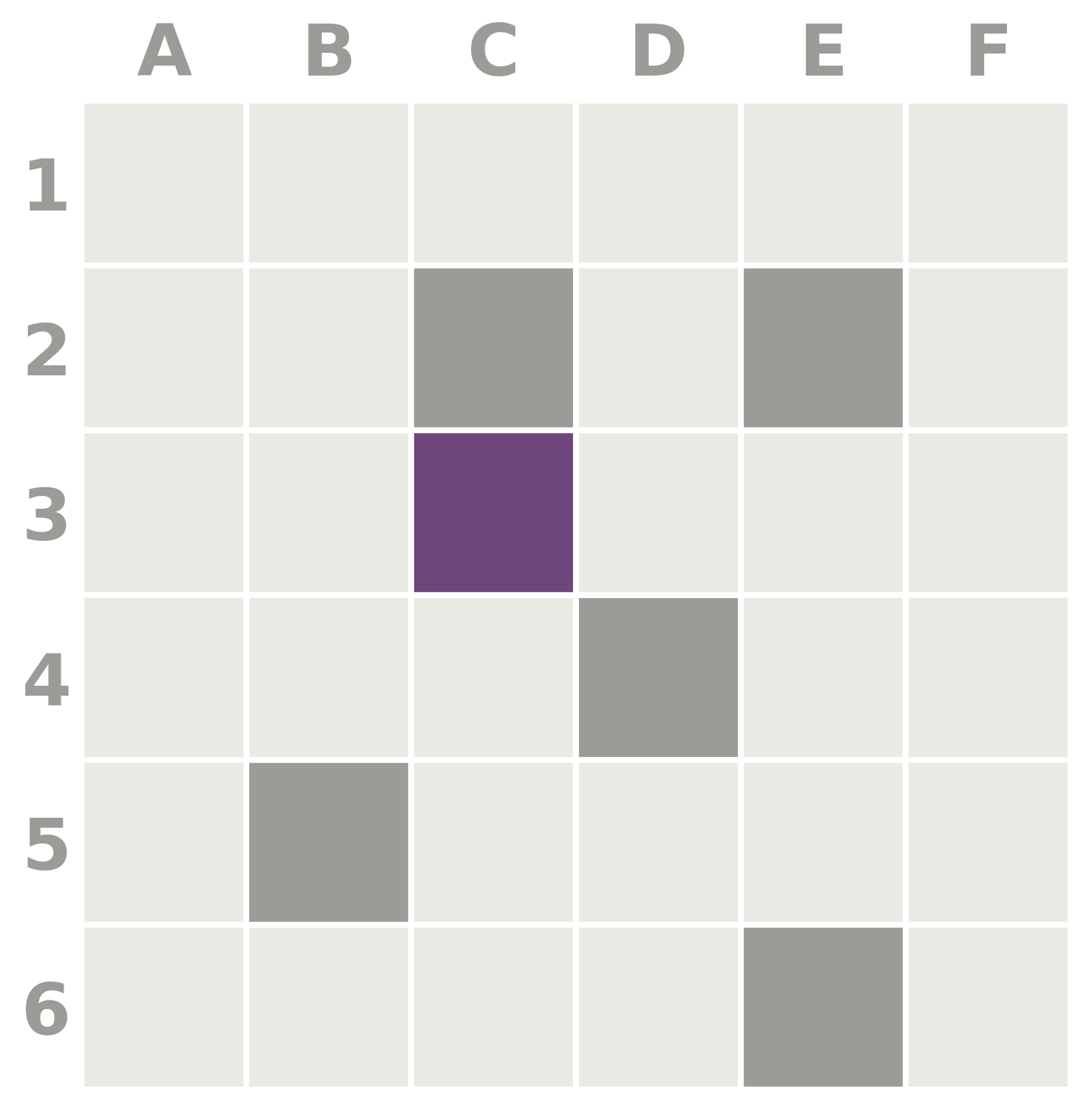}}} & \multirow[]{2}{*}{Human} & $\star$ & What is the location of one red tile? & \texttt{(topleft (coloredTiles Red))} & 4.58 \\
 &  & $\epsdice{6}$ & How many tiles is the purple ship? & \texttt{(size Purple)} & 1.58 \\
\cline{2-6}
 & \multirow[]{2}{*}{CodeLlama} & $\star$ & What is the location of one red tile? & \texttt{(topleft (coloredTiles Red))} & 4.58 \\
 &  & $\epsdice{6}$ & How many tiles is the red ship? & \texttt{(size Red)} & 1.41 \\
\cline{2-6}
 & \multirow[]{2}{*}{GPT-4} & $\star$ & Where is one blue tile located? & \texttt{(topleft (coloredTiles Blue))} & 4.58 \\
 &  & $\epsdice{6}$ & How many tiles is the purple ship? & \texttt{(size Purple)} & 1.58 \\
\cline{2-6}
 & \multirow[]{2}{*}{Grammar} & $\star$ & --- & \texttt{(bottomright (union (intersection (set AllTiles) (coloredTiles Red)) (intersection (unique (set...} & 4.65 \\
 &  & $\epsdice{6}$ & --- & \texttt{(color 3F)} & 1.43 \\
\cline{1-6}
\multirow[]{8}{*}{\shortstack[m]{Trial 3 \\ \includegraphics[width=1in]{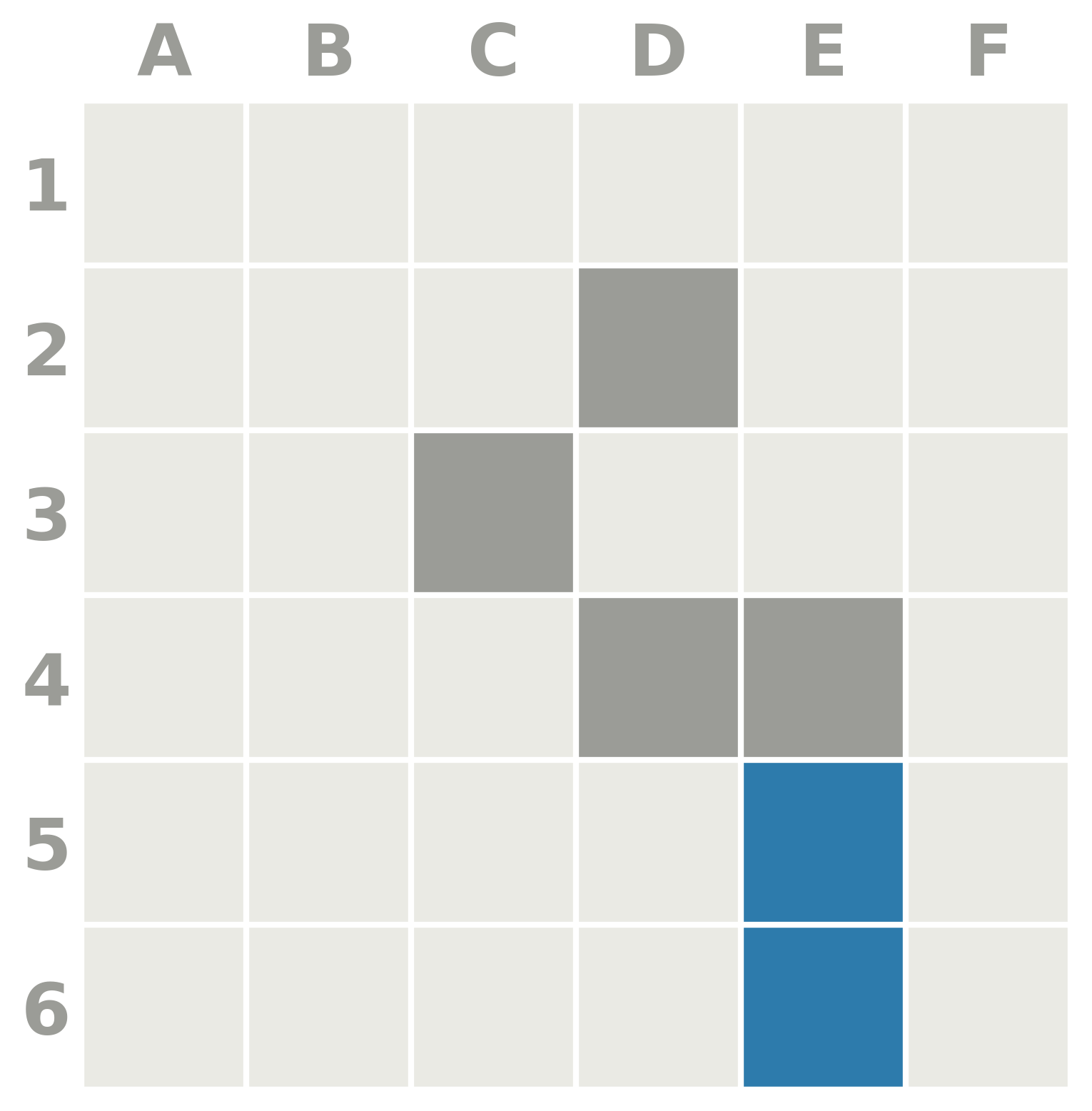}}} & \multirow[]{2}{*}{Human} & $\star$ & At what location is the top left part of the purple ship? & \texttt{(topleft (coloredTiles Purple))} & 4.62 \\
 &  & $\epsdice{6}$ & How many tiles is the red ship? & \texttt{(size Red)} & 1.44 \\
\cline{2-6}
 & \multirow[]{2}{*}{CodeLlama} & $\star$ & How many tiles is the purple ship? & \texttt{(size Purple)} & 1.44 \\
 &  & $\epsdice{6}$ & Is the red ship horizontal? & \texttt{(== (orient Red) H)} & 0.99 \\
\cline{2-6}
 & \multirow[]{2}{*}{GPT-4} & $\star$ & How many tiles is the red ship? & \texttt{(size Red)} & 1.44 \\
 &  & $\epsdice{6}$ & Is the red ship horizontal? & \texttt{(== (orient Red) H)} & 0.99 \\
\cline{2-6}
 & \multirow[]{2}{*}{Grammar} & $\star$ & --- & \texttt{(bottomright (unique (intersection (set AllTiles) (coloredTiles Purple))))} & 4.73 \\
 &  & $\epsdice{6}$ & --- & \texttt{(orient Purple)} & 0.99 \\
\cline{1-6}
\multirow[]{8}{*}{\shortstack[m]{Trial 4 \\ \includegraphics[width=1in]{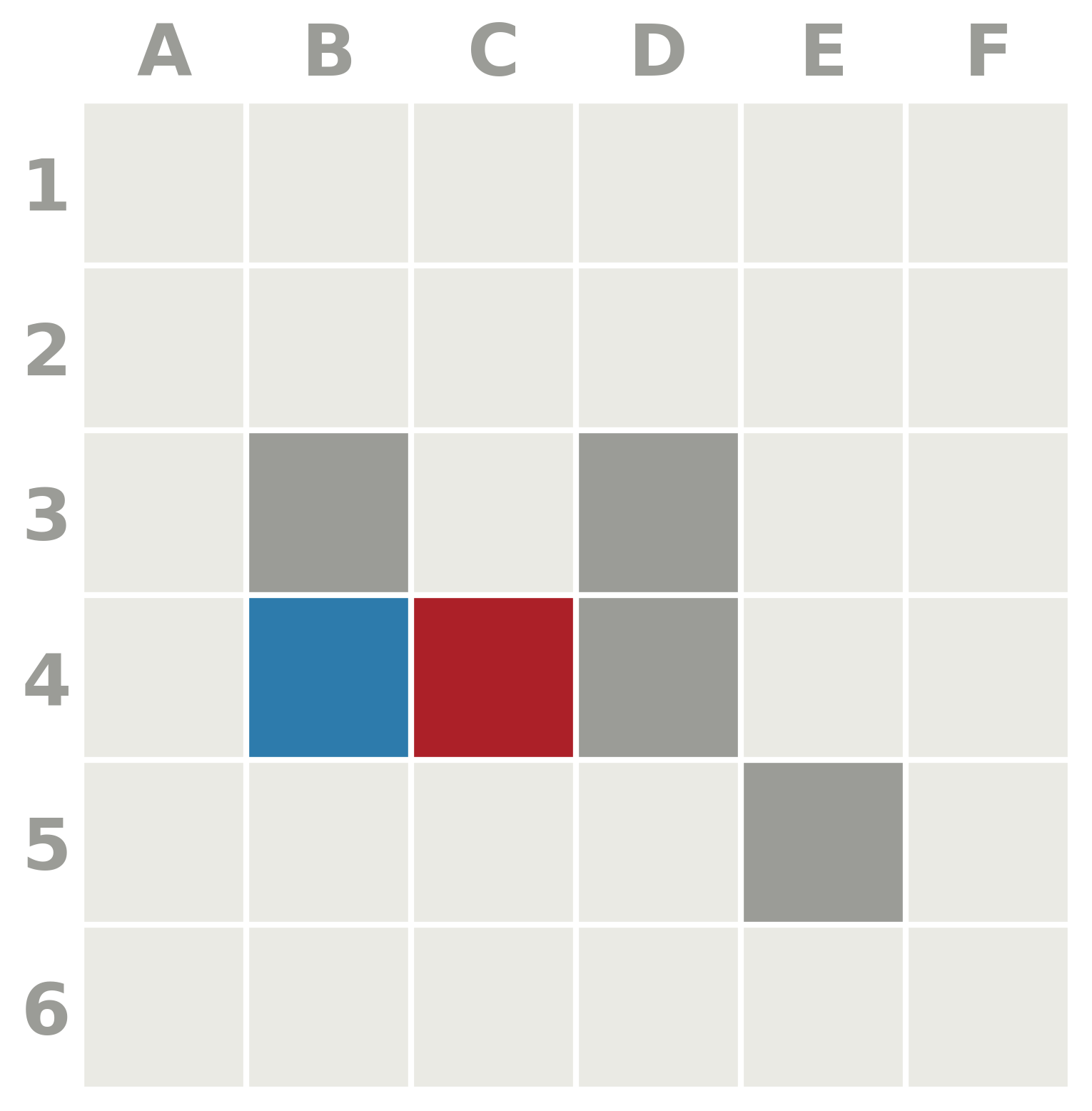}}} & \multirow[]{2}{*}{Human} & $\star$ & At what location is the top left part of the purple ship? & \texttt{(topleft (coloredTiles Purple))} & 4.62 \\
 &  & $\epsdice{6}$ & At what location is the top left part of the purple ship? & \texttt{(topleft (coloredTiles Purple))} & 4.62 \\
\cline{2-6}
 & \multirow[]{2}{*}{CodeLlama} & $\star$ & How many tiles is the red ship? & \texttt{(size Red)} & 1.57 \\
 &  & $\epsdice{6}$ & How many tiles is the red ship? & \texttt{(size Red)} & 1.57 \\
\cline{2-6}
 & \multirow[]{2}{*}{GPT-4} & $\star$ & How many tiles is the red ship? & \texttt{(size Red)} & 1.57 \\
 &  & $\epsdice{6}$ & How many tiles is the red ship? & \texttt{(size Red)} & 1.57 \\
\cline{2-6}
 & \multirow[]{2}{*}{Grammar} & $\star$ & --- & \texttt{(bottomright (intersection (set AllTiles) (intersection (coloredTiles Purple) (set AllTiles))))} & 4.64 \\
 &  & $\epsdice{6}$ & --- & \texttt{(== (orient Blue) H)} & 0.91 \\
\cline{1-6}
\multirow[]{8}{*}{\shortstack[m]{Trial 5 \\ \includegraphics[width=1in]{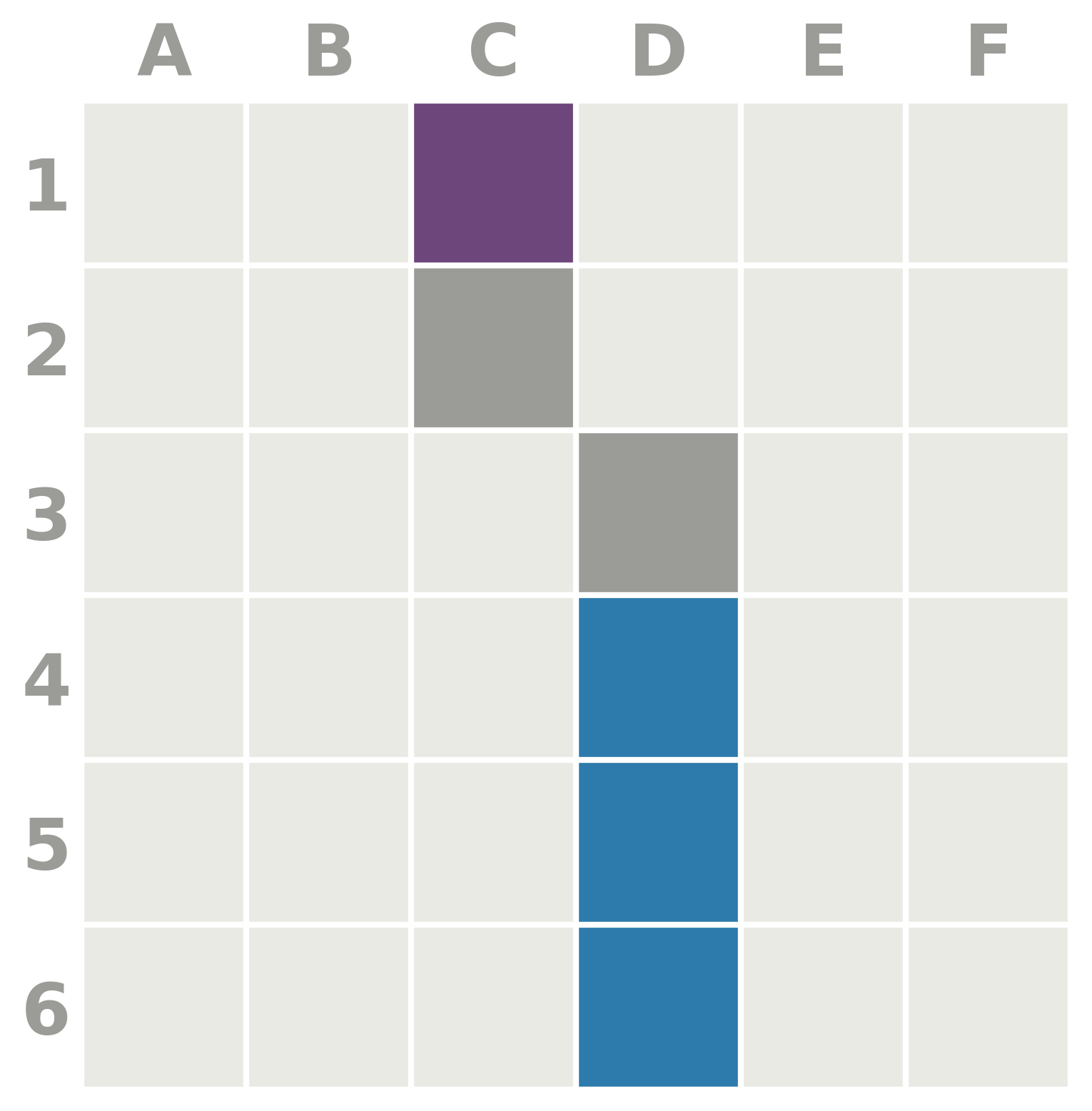}}} & \multirow[]{2}{*}{Human} & $\star$ & At what location is the top left part of the red ship? & \texttt{(topleft (coloredTiles Red))} & 4.66 \\
 &  & $\epsdice{6}$ & How many tiles is the purple ship? & \texttt{(size Purple)} & 1.57 \\
\cline{2-6}
 & \multirow[]{2}{*}{CodeLlama} & $\star$ & At what location is the bottom right part of the purple ship? & \texttt{(bottomright (coloredTiles Purple))} & 1.90 \\
 &  & $\epsdice{6}$ & How many tiles is the purple ship? & \texttt{(size Purple)} & 1.57 \\
\cline{2-6}
 & \multirow[]{2}{*}{GPT-4} & $\star$ & How many tiles is the purple ship? & \texttt{(size Purple)} & 1.57 \\
 &  & $\epsdice{6}$ & How many tiles is the purple ship? & \texttt{(size Purple)} & 1.57 \\
\cline{2-6}
 & \multirow[]{2}{*}{Grammar} & $\star$ & --- & \texttt{(topleft (unique (coloredTiles Red)))} & 4.66 \\
 &  & $\epsdice{6}$ & --- & \texttt{(size Purple)} & 1.57 \\
\cline{1-6}
\multirow[]{8}{*}{\shortstack[m]{Trial 6 \\ \includegraphics[width=1in]{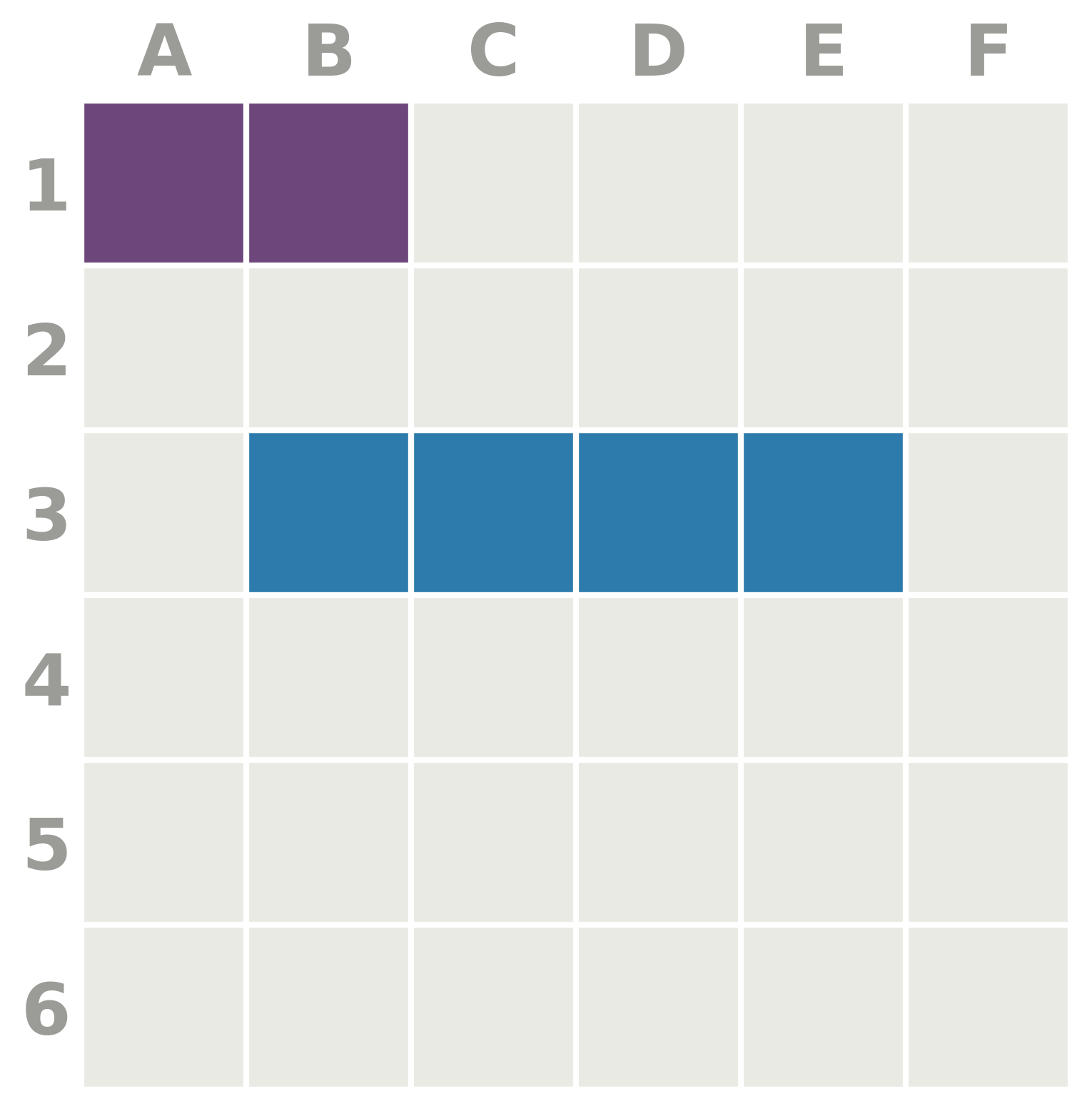}}} & \multirow[]{2}{*}{Human} & $\star$ & At what location is the top left part of the red ship? & \texttt{(topleft (coloredTiles Red))} & 4.73 \\
 &  & $\epsdice{6}$ & Does the red ship touch both other ships? & \texttt{(and (touch Red Blue) (touch Red Purple))} & 0.60 \\
\cline{2-6}
 & \multirow[]{2}{*}{CodeLlama} & $\star$ & At what location is the top left part of the red ship? & \texttt{(topleft (coloredTiles Red))} & 4.73 \\
 &  & $\epsdice{6}$ & Is the red ship 2 tiles long? & \texttt{(== (size Red) 2)} & 1.00 \\
\cline{2-6}
 & \multirow[]{2}{*}{GPT-4} & $\star$ & Where is a tile of the red ship? & \texttt{(topleft (coloredTiles Red))} & 4.73 \\
 &  & $\epsdice{6}$ & How many tiles is the red ship? & \texttt{(size Red)} & 1.50 \\
\cline{2-6}
 & \multirow[]{2}{*}{Grammar} & $\star$ & --- & \texttt{(topleft (intersection (set AllTiles) (coloredTiles Red)))} & 4.73 \\
 &  & $\epsdice{6}$ & --- & \texttt{(+ (== (color 4A) Red) TRUE)} & 0.52 \\
\cline{1-6}
\multirow[]{8}{*}{\shortstack[m]{Trial 7 \\ \includegraphics[width=1in]{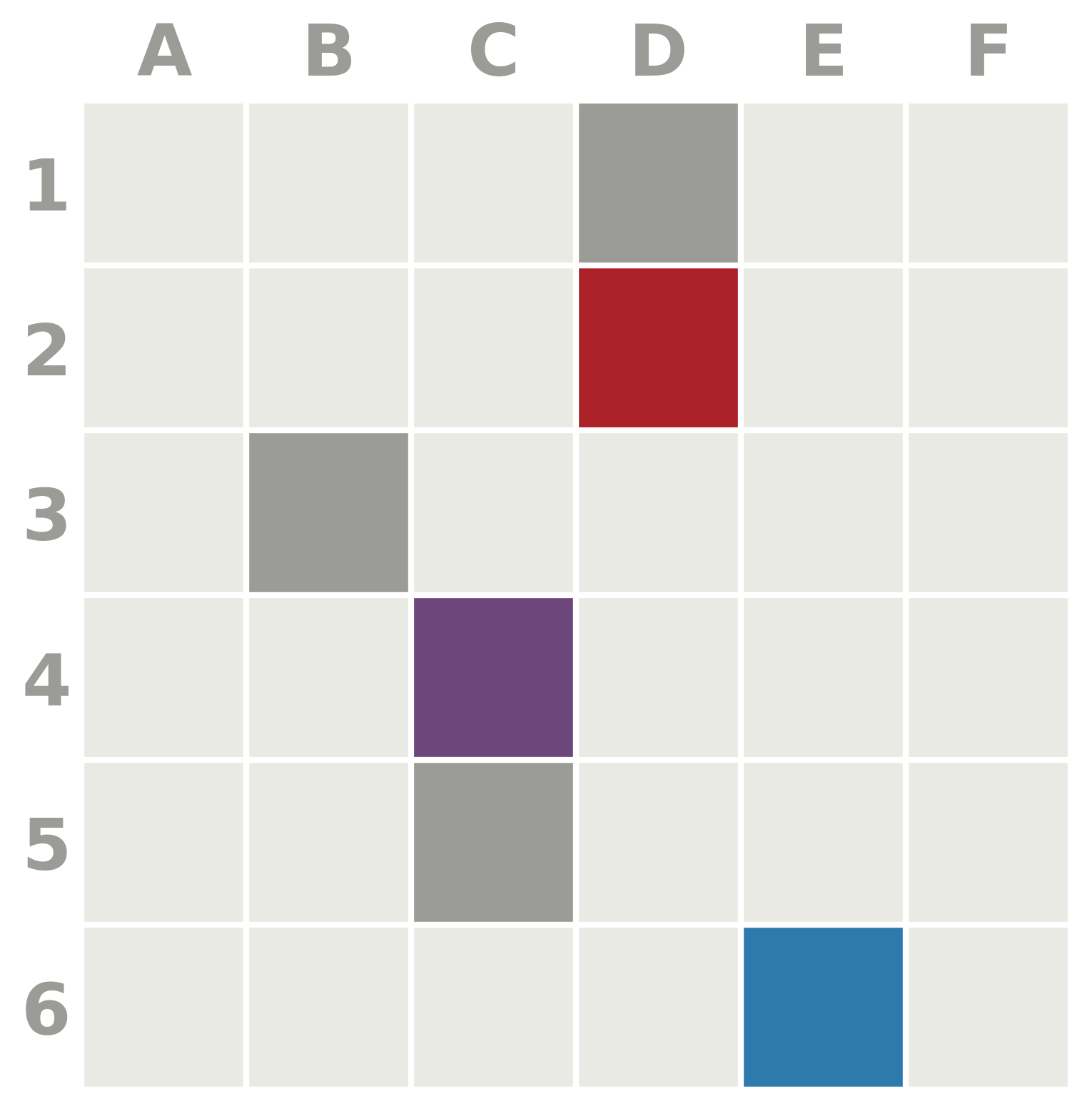}}} & \multirow[]{2}{*}{Human} & $\star$ & How many tiles are occupied by ships? & \texttt{(++ (map (lambda x0 (size x0)) (set AllColors)))} & 2.48 \\
 &  & $\epsdice{6}$ & Is there a blue tile at 5E? & \texttt{(== (color 5E) Blue)} & 0.87 \\
\cline{2-6}
 & \multirow[]{2}{*}{CodeLlama} & $\star$ & What is the location of one red tile? & \texttt{(topleft (coloredTiles Red))} & 1.79 \\
 &  & $\epsdice{6}$ & How many tiles is the red ship? & \texttt{(size Red)} & 1.58 \\
\cline{2-6}
 & \multirow[]{2}{*}{GPT-4} & $\star$ & How many tiles is the red ship? & \texttt{(size Red)} & 1.58 \\
 &  & $\epsdice{6}$ & How many tiles is the red ship? & \texttt{(size Red)} & 1.58 \\
\cline{2-6}
 & \multirow[]{2}{*}{Grammar} & $\star$ & --- & \texttt{(- (setSize (union (coloredTiles Red) (setDifference (union (unique (union (intersection...} & 3.28 \\
 &  & $\epsdice{6}$ & --- & \texttt{(== (color 6D) Blue)} & 0.98 \\
\cline{1-6}
\multirow[]{8}{*}{\shortstack[m]{Trial 8 \\ \includegraphics[width=1in]{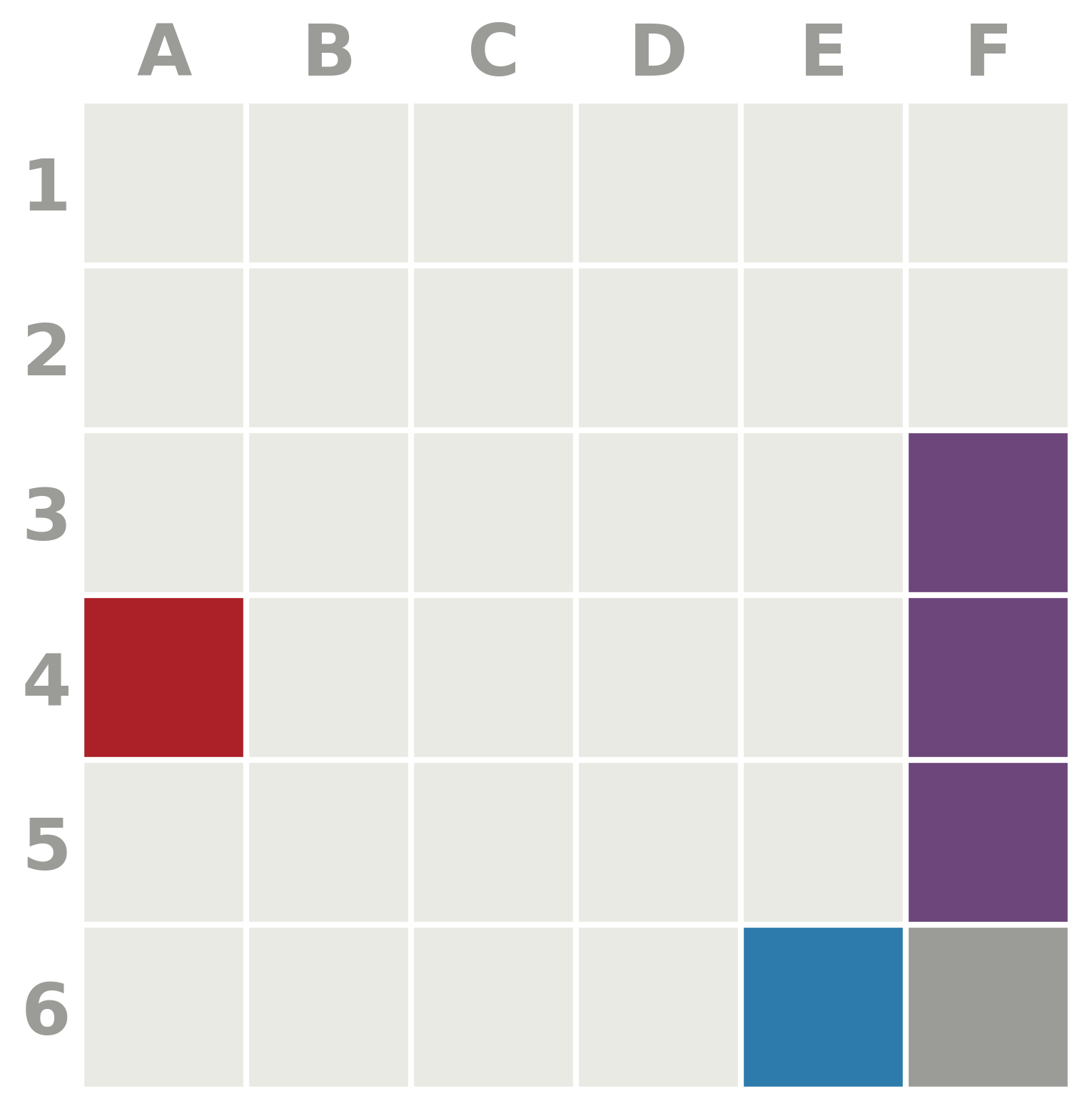}}} & \multirow[]{2}{*}{Human} & $\star$ & At what location is the bottom right part of the red ship? & \texttt{(bottomright (coloredTiles Red))} & 2.41 \\
 &  & $\epsdice{6}$ & Is the red ship horizontal? & \texttt{(== (orient Red) H)} & 0.85 \\
\cline{2-6}
 & \multirow[]{2}{*}{CodeLlama} & $\star$ & What is the location of one blue tile? & \texttt{(topleft (coloredTiles Blue))} & 2.58 \\
 &  & $\epsdice{6}$ & At what location is the bottom right part of the red ship? & \texttt{(bottomright (coloredTiles Red))} & 2.41 \\
\cline{2-6}
 & \multirow[]{2}{*}{GPT-4} & $\star$ & How many tiles is the blue ship? & \texttt{(size Blue)} & 1.58 \\
 &  & $\epsdice{6}$ & How many tiles is the red ship? & \texttt{(size Red)} & 1.57 \\
\cline{2-6}
 & \multirow[]{2}{*}{Grammar} & $\star$ & --- & \texttt{(- (setSize (coloredTiles Blue)) (setSize (setDifference (intersection (coloredTiles (color 3A))...} & 3.16 \\
 &  & $\epsdice{6}$ & --- & \texttt{(orient Red)} & 0.85 \\
\cline{1-6}
\multirow[]{8}{*}{\shortstack[m]{Trial 9 \\ \includegraphics[width=1in]{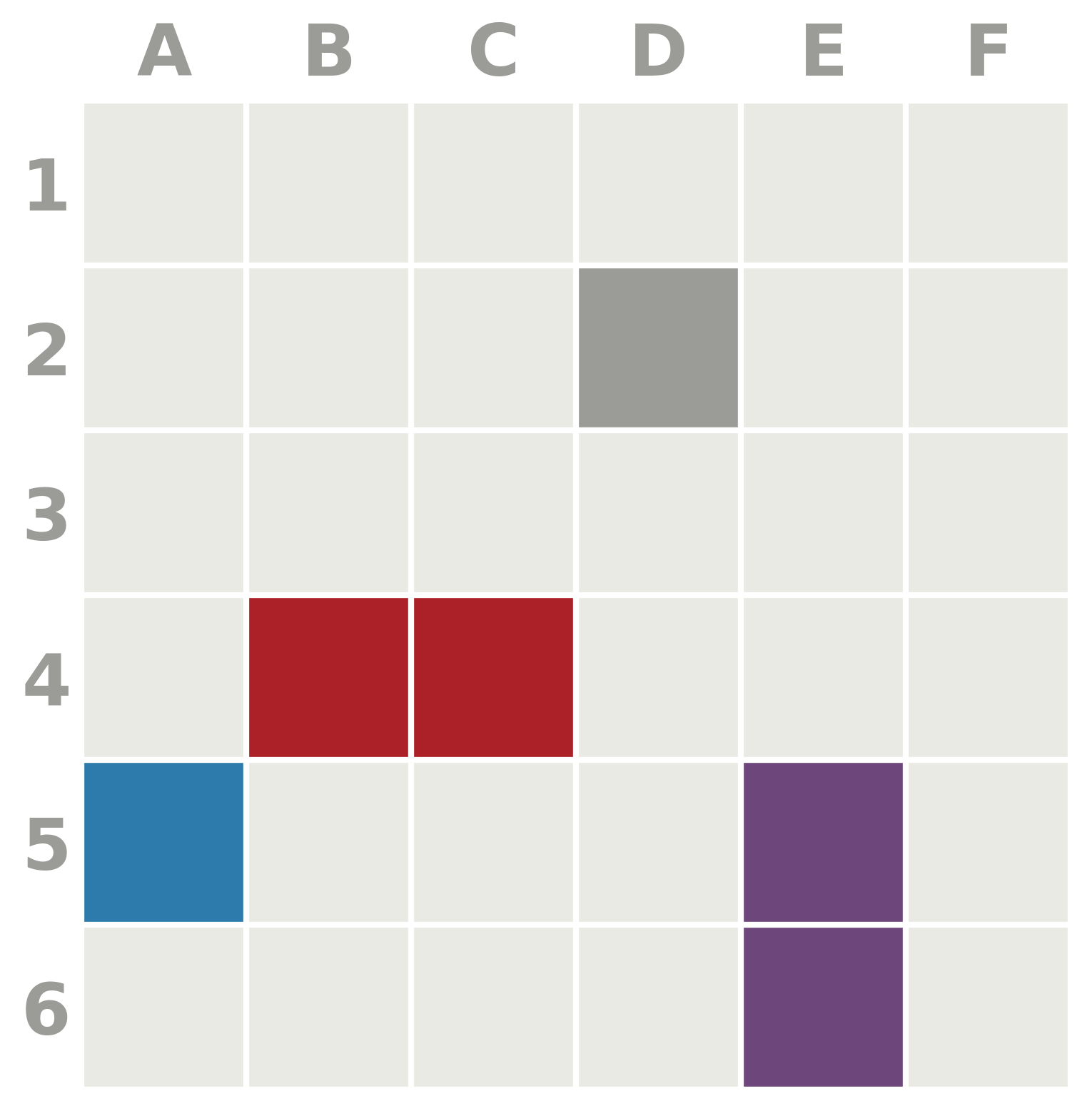}}} & \multirow[]{2}{*}{Human} & $\star$ & How many tiles in row 4 are occupied by ships? & \texttt{(++ (map (lambda x0 (++ (map (lambda y0 (== (rowL y0) 4)) (coloredTiles x0)))) (set AllColors)))} & 1.73 \\
 &  & $\epsdice{6}$ & Is the blue ship 3 tiles long? & \texttt{(== (size Blue) 3)} & 0.89 \\
\cline{2-6}
 & \multirow[]{2}{*}{CodeLlama} & $\star$ & Where is the bottom right tile of the blue ship? & \texttt{(bottomright (coloredTiles Blue))} & 2.25 \\
 &  & $\epsdice{6}$ & How many tiles is the red ship? & \texttt{(size Red)} & 1.54 \\
\cline{2-6}
 & \multirow[]{2}{*}{GPT-4} & $\star$ & How many tiles is the blue ship? & \texttt{(size Blue)} & 1.58 \\
 &  & $\epsdice{6}$ & How many tiles is the red ship? & \texttt{(size Red)} & 1.54 \\
\cline{2-6}
 & \multirow[]{2}{*}{Grammar} & $\star$ & --- & \texttt{(- (setSize (coloredTiles Blue)) (setSize (setDifference (intersection (coloredTiles (color 3A))...} & 3.21 \\
 &  & $\epsdice{6}$ & --- & \texttt{(== (color 5B) Water)} & 0.99 \\
\cline{1-6}
\multirow[]{8}{*}{\shortstack[m]{Trial 10 \\ \includegraphics[width=1in]{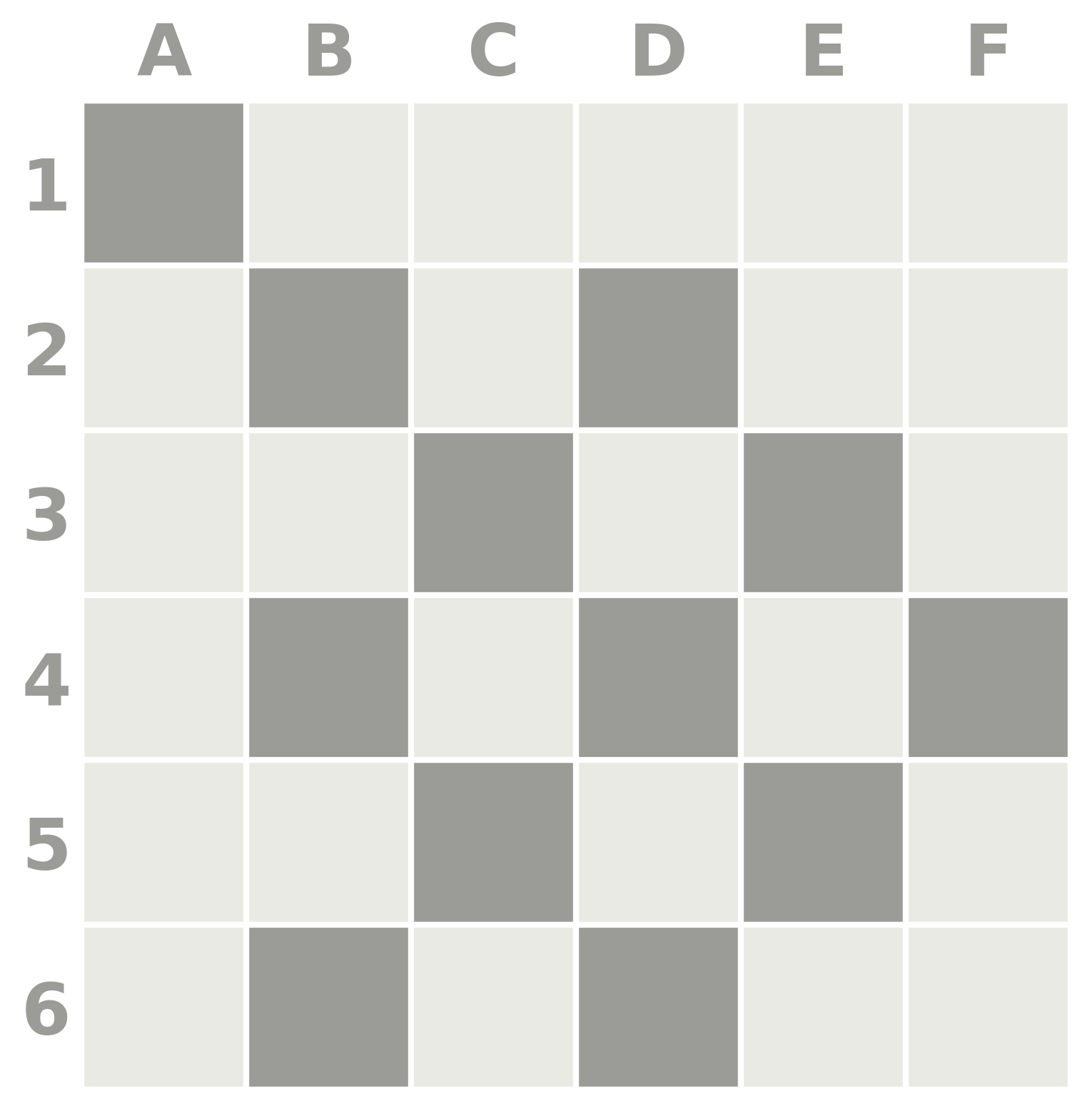}}} & \multirow[]{2}{*}{Human} & $\star$ & What is the location of one blue tile? & \texttt{(topleft (coloredTiles Blue))} & 3.64 \\
 &  & $\epsdice{6}$ & How many tiles is the blue ship? & \texttt{(size Blue)} & 1.12 \\
\cline{2-6}
 & \multirow[]{2}{*}{CodeLlama} & $\star$ & Where is the bottom right part of the purple ship? & \texttt{(bottomright (coloredTiles Purple))} & 3.80 \\
 &  & $\epsdice{6}$ & What is the top left tile of the blue ship? & \texttt{(topleft (coloredTiles Blue))} & 3.64 \\
\cline{2-6}
 & \multirow[]{2}{*}{GPT-4} & $\star$ & Where is the purple ship located? & \texttt{(topleft (coloredTiles Purple))} & 3.64 \\
 &  & $\epsdice{6}$ & How many tiles is the red ship? & \texttt{(size Red)} & 1.12 \\
\cline{2-6}
 & \multirow[]{2}{*}{Grammar} & $\star$ & --- & \texttt{(bottomright (setDifference (coloredTiles Blue) (coloredTiles (color 6E))))} & 3.93 \\
 &  & $\epsdice{6}$ & --- & \texttt{(color 1C)} & 1.95 \\
\cline{1-6}
\multirow[]{8}{*}{\shortstack[m]{Trial 11 \\ \includegraphics[width=1in]{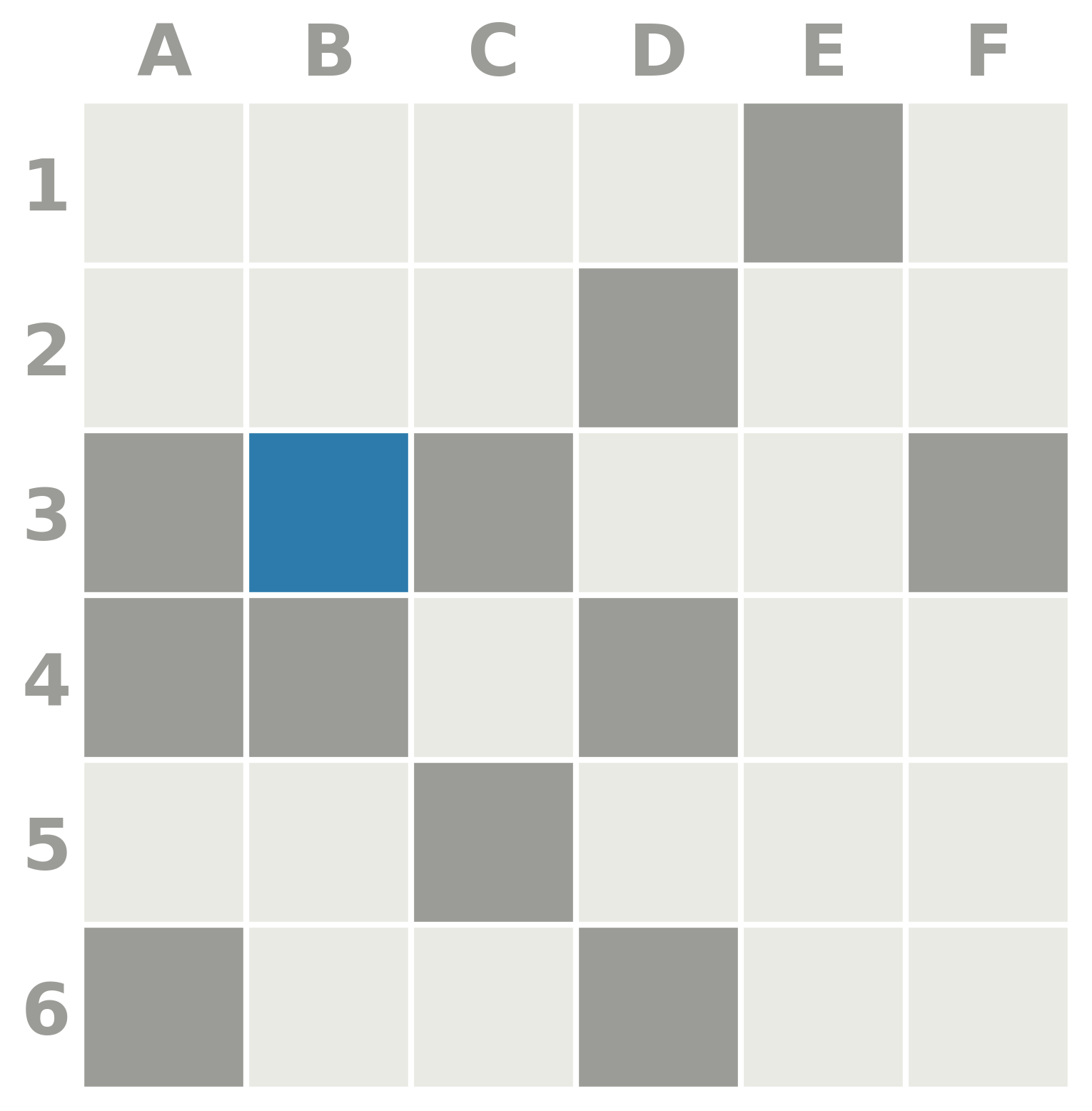}}} & \multirow[]{2}{*}{Human} & $\star$ & What is the location of one purple tile? & \texttt{(topleft (coloredTiles Purple))} & 3.88 \\
 &  & $\epsdice{6}$ & How many tiles is the blue ship? & \texttt{(size Blue)} & 0.97 \\
\cline{2-6}
 & \multirow[]{2}{*}{CodeLlama} & $\star$ & What is the location of one purple tile? & \texttt{(topleft (coloredTiles Purple))} & 3.88 \\
 &  & $\epsdice{6}$ & Is there a ship at 2E? & \texttt{(color 2E)} & 1.07 \\
\cline{2-6}
 & \multirow[]{2}{*}{GPT-4} & $\star$ & What is the position of one tile of the red ship? & \texttt{(topleft (coloredTiles Red))} & 3.88 \\
 &  & $\epsdice{6}$ & How many tiles is the blue ship? & \texttt{(size Blue)} & 0.97 \\
\cline{2-6}
 & \multirow[]{2}{*}{Grammar} & $\star$ & --- & \texttt{(topleft (coloredTiles Purple))} & 3.88 \\
 &  & $\epsdice{6}$ & --- & \texttt{(+ (size Red) (rowL 3F))} & 1.02 \\
\pagebreak 
\multirow[]{8}{*}{\shortstack[m]{Trial 12 \\ \includegraphics[width=1in]{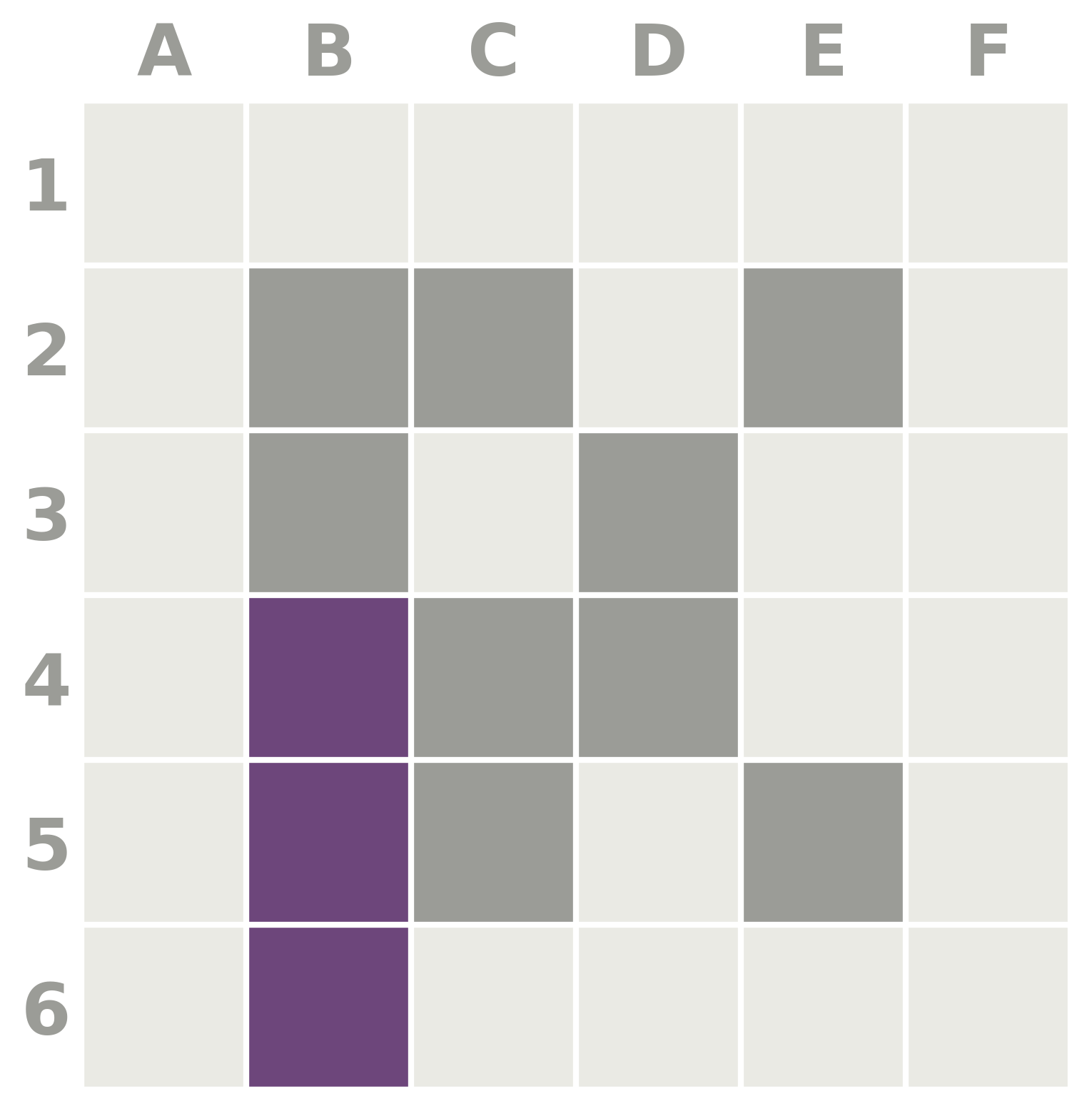}}} & \multirow[]{2}{*}{Human} & $\star$ & At what location is the top left part of the blue ship? & \texttt{(topleft (coloredTiles Blue))} & 4.16 \\
 &  & $\epsdice{6}$ & Is there any part of the blue ship in row 1? & \texttt{(any (map (lambda y0 (== (rowL y0) 1)) (coloredTiles Blue)))} & 0.97 \\
\cline{2-6}
 & \multirow[]{2}{*}{CodeLlama} & $\star$ & At what location is the top left part of the blue ship? & \texttt{(topleft (coloredTiles Blue))} & 4.16 \\
 &  & $\epsdice{6}$ & How many tiles is the blue ship? & \texttt{(size Blue)} & 1.47 \\
\cline{2-6}
 & \multirow[]{2}{*}{GPT-4} & $\star$ & How many tiles is the blue ship? & \texttt{(size Blue)} & 1.47 \\
 &  & $\epsdice{6}$ & How many tiles is the blue ship? & \texttt{(size Blue)} & 1.47 \\
\cline{2-6}
 & \multirow[]{2}{*}{Grammar} & $\star$ & --- & \texttt{(topleft (unique (coloredTiles Red)))} & 4.16 \\
 &  & $\epsdice{6}$ & --- & \texttt{(color 6D)} & 1.13 \\
\cline{1-6}
\multirow[]{8}{*}{\shortstack[m]{Trial 13 \\ \includegraphics[width=1in]{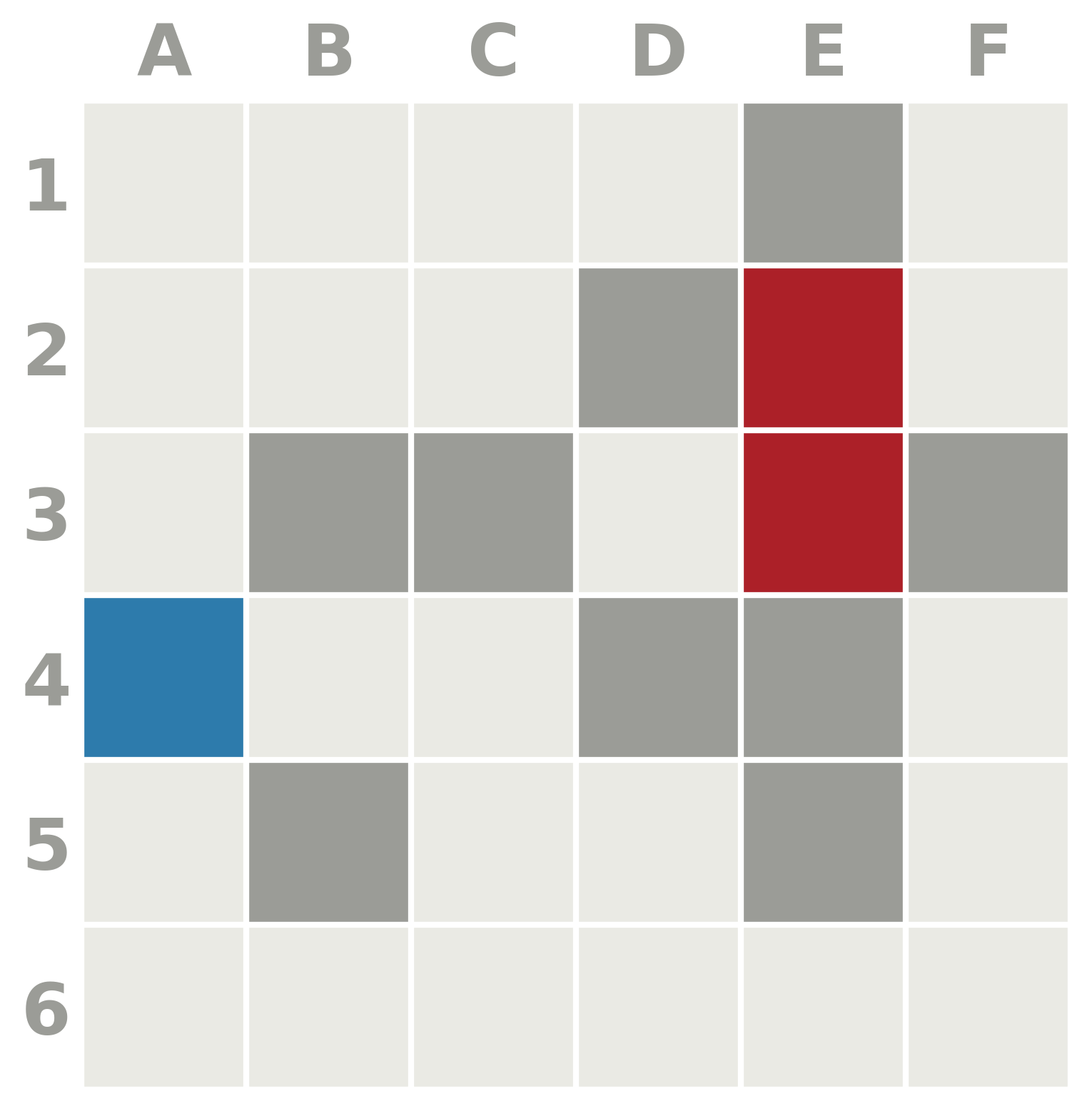}}} & \multirow[]{2}{*}{Human} & $\star$ & At what location is the top left part of the purple ship? & \texttt{(topleft (coloredTiles Purple))} & 3.99 \\
 &  & $\epsdice{6}$ & How many tiles is the blue ship? & \texttt{(size Blue)} & 1.57 \\
\cline{2-6}
 & \multirow[]{2}{*}{CodeLlama} & $\star$ & At what location is the top left part of the purple ship? & \texttt{(topleft (coloredTiles Purple))} & 3.99 \\
 &  & $\epsdice{6}$ & How many tiles is the blue ship? & \texttt{(size Blue)} & 1.57 \\
\cline{2-6}
 & \multirow[]{2}{*}{GPT-4} & $\star$ & How many tiles is the blue ship? & \texttt{(size Blue)} & 1.57 \\
 &  & $\epsdice{6}$ & How many tiles is the blue ship? & \texttt{(size Blue)} & 1.57 \\
\cline{2-6}
 & \multirow[]{2}{*}{Grammar} & $\star$ & --- & \texttt{(topleft (coloredTiles Purple))} & 3.99 \\
 &  & $\epsdice{6}$ & --- & \texttt{(setSize (setDifference (coloredTiles (color 1E)) (unique (coloredTiles Blue))))} & 2.03 \\
\cline{1-6}
\multirow[]{8}{*}{\shortstack[m]{Trial 14 \\ \includegraphics[width=1in]{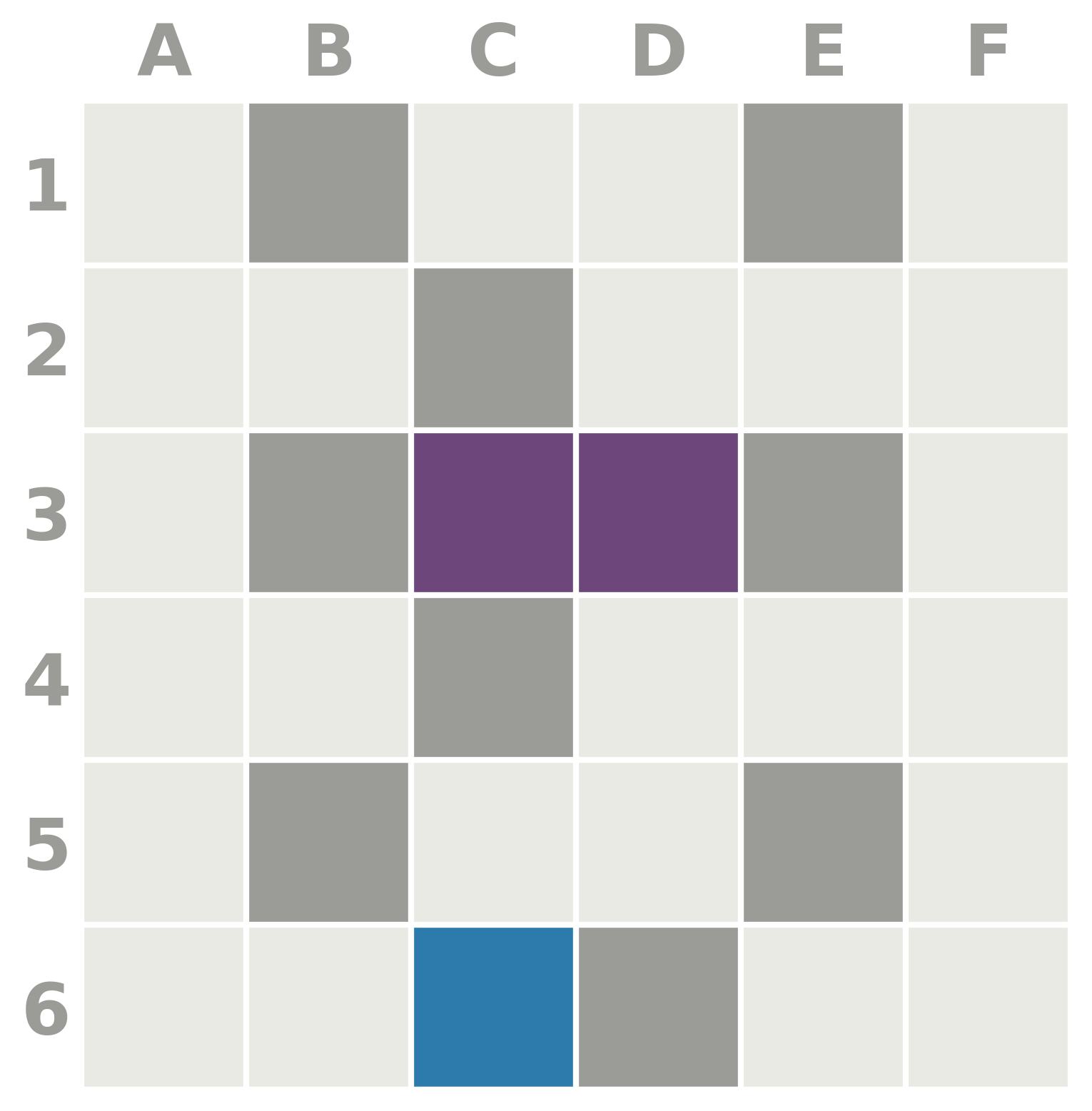}}} & \multirow[]{2}{*}{Human} & $\star$ & What is the location of one red tile? & \texttt{(topleft (coloredTiles Red))} & 4.00 \\
 &  & $\epsdice{6}$ & Is there any part of the red ship in column A? & \texttt{(any (map (lambda y0 (== (colL y0) 1)) (coloredTiles Red)))} & 0.95 \\
\cline{2-6}
 & \multirow[]{2}{*}{CodeLlama} & $\star$ & At what location is the top left part of the blue ship? & \texttt{(topleft (coloredTiles Blue))} & 1.58 \\
 &  & $\epsdice{6}$ & How many tiles is the red ship? & \texttt{(size Red)} & 1.39 \\
\cline{2-6}
 & \multirow[]{2}{*}{GPT-4} & $\star$ & How many tiles is the red ship? & \texttt{(size Red)} & 1.39 \\
 &  & $\epsdice{6}$ & Is the blue ship vertical? & \texttt{(== (orient Blue) V)} & 0.93 \\
\cline{2-6}
 & \multirow[]{2}{*}{Grammar} & $\star$ & --- & \texttt{(topleft (coloredTiles Red))} & 4.00 \\
 &  & $\epsdice{6}$ & --- & \texttt{(topleft (setDifference (unique (union (coloredTiles Water) (set AllTiles))) (unique (union...} & 2.71 \\
\cline{1-6}
\multirow[]{8}{*}{\shortstack[m]{Trial 15 \\ \includegraphics[width=1in]{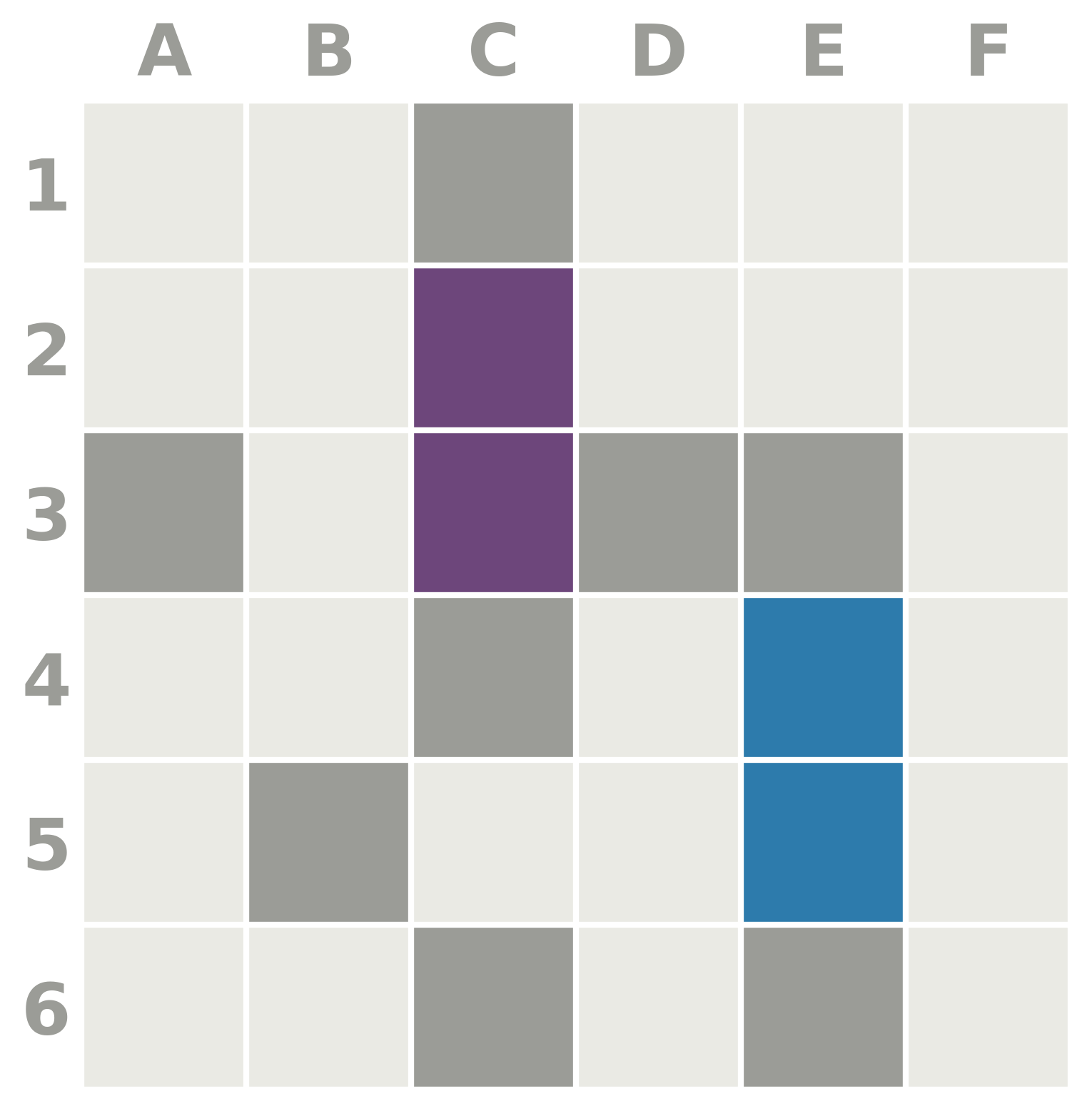}}} & \multirow[]{2}{*}{Human} & $\star$ & At what location is the top left part of the red ship? & \texttt{(topleft (coloredTiles Red))} & 4.18 \\
 &  & $\epsdice{6}$ & Is the red ship 3 tiles long? & \texttt{(== (size Red) 3)} & 0.83 \\
\cline{2-6}
 & \multirow[]{2}{*}{CodeLlama} & $\star$ & How many tiles is the red ship? & \texttt{(size Red)} & 1.27 \\
 &  & $\epsdice{6}$ & How many tiles is the red ship? & \texttt{(size Red)} & 1.27 \\
\cline{2-6}
 & \multirow[]{2}{*}{GPT-4} & $\star$ & How many tiles is the red ship? & \texttt{(size Red)} & 1.27 \\
 &  & $\epsdice{6}$ & How many tiles is the blue ship? & \texttt{(size Blue)} & 0.00 \\
\cline{2-6}
 & \multirow[]{2}{*}{Grammar} & $\star$ & --- & \texttt{(topleft (coloredTiles Red))} & 4.18 \\
 &  & $\epsdice{6}$ & --- & \texttt{(- (setSize (coloredTiles Water)) (colL 2B))} & 1.27 \\
\multirow[]{8}{*}{\shortstack[m]{Trial 16 \\ \includegraphics[width=1in]{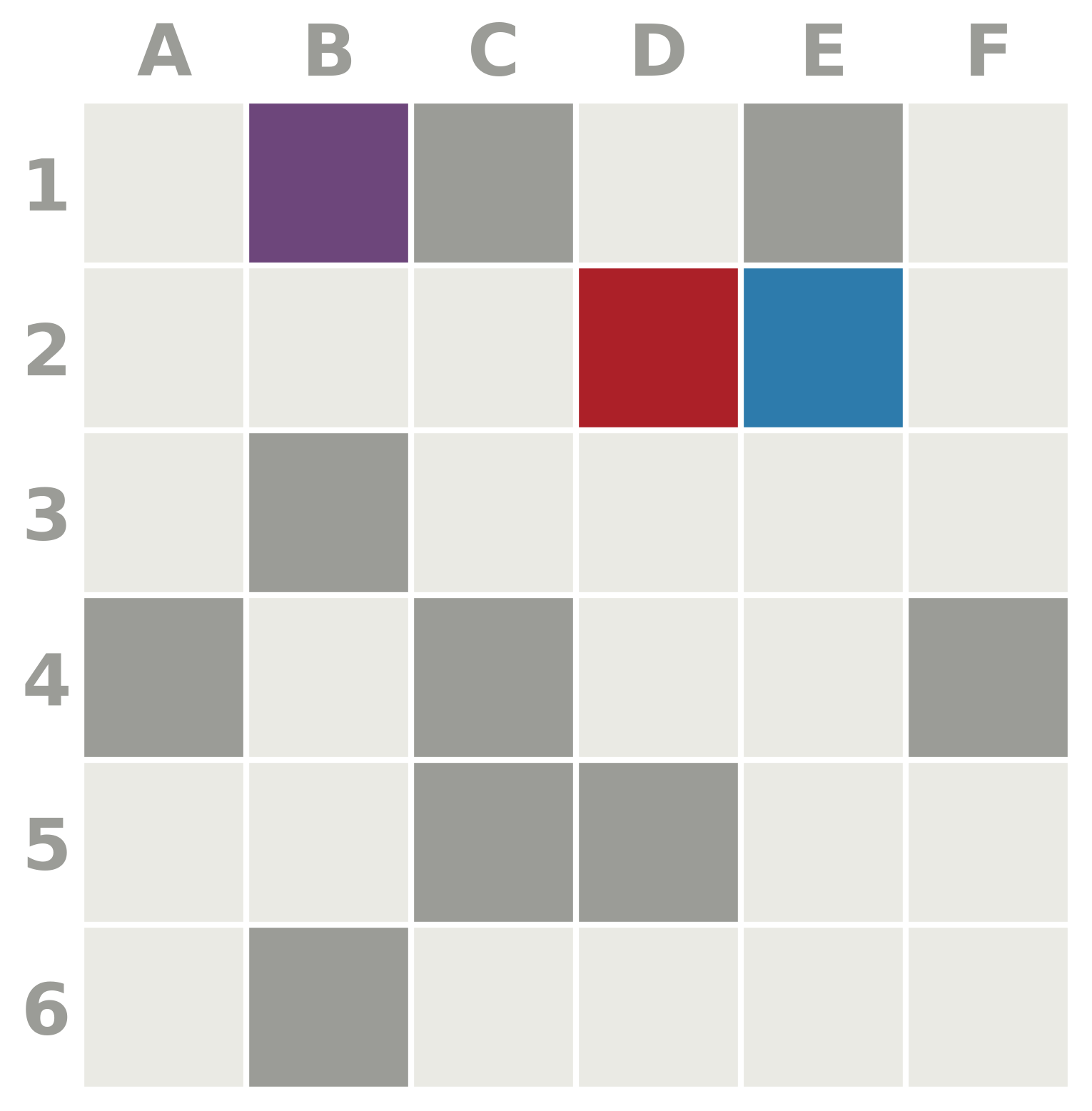}}} & \multirow[]{2}{*}{Human} & $\star$ & How many tiles in row 2 are occupied by ships? & \texttt{(++ (map (lambda x0 (++ (map (lambda y0 (== (rowL y0) 2)) (coloredTiles x0)))) (set AllColors)))} & 1.93 \\
 &  & $\epsdice{6}$ & Is the red ship horizontal? & \texttt{(== (orient Red) H)} & 0.86 \\
\cline{2-6}
 & \multirow[]{2}{*}{CodeLlama} & $\star$ & What is the location of one red tile? & \texttt{(topleft (coloredTiles Red))} & 1.99 \\
 &  & $\epsdice{6}$ & How many tiles is the red ship? & \texttt{(size Red)} & 1.53 \\
\cline{2-6}
 & \multirow[]{2}{*}{GPT-4} & $\star$ & How many tiles is the red ship? & \texttt{(size Red)} & 1.53 \\
 &  & $\epsdice{6}$ & How many tiles is the red ship? & \texttt{(size Red)} & 1.53 \\
\cline{2-6}
 & \multirow[]{2}{*}{Grammar} & $\star$ & --- & \texttt{(- (setSize (coloredTiles Blue)) (setSize (setDifference (intersection (coloredTiles (color 3A))...} & 3.13 \\
 &  & $\epsdice{6}$ & --- & \texttt{(setSize (coloredTiles (color 3B)))} & 2.14 \\
\cline{1-6}
\multirow[]{8}{*}{\shortstack[m]{Trial 17 \\ \includegraphics[width=1in]{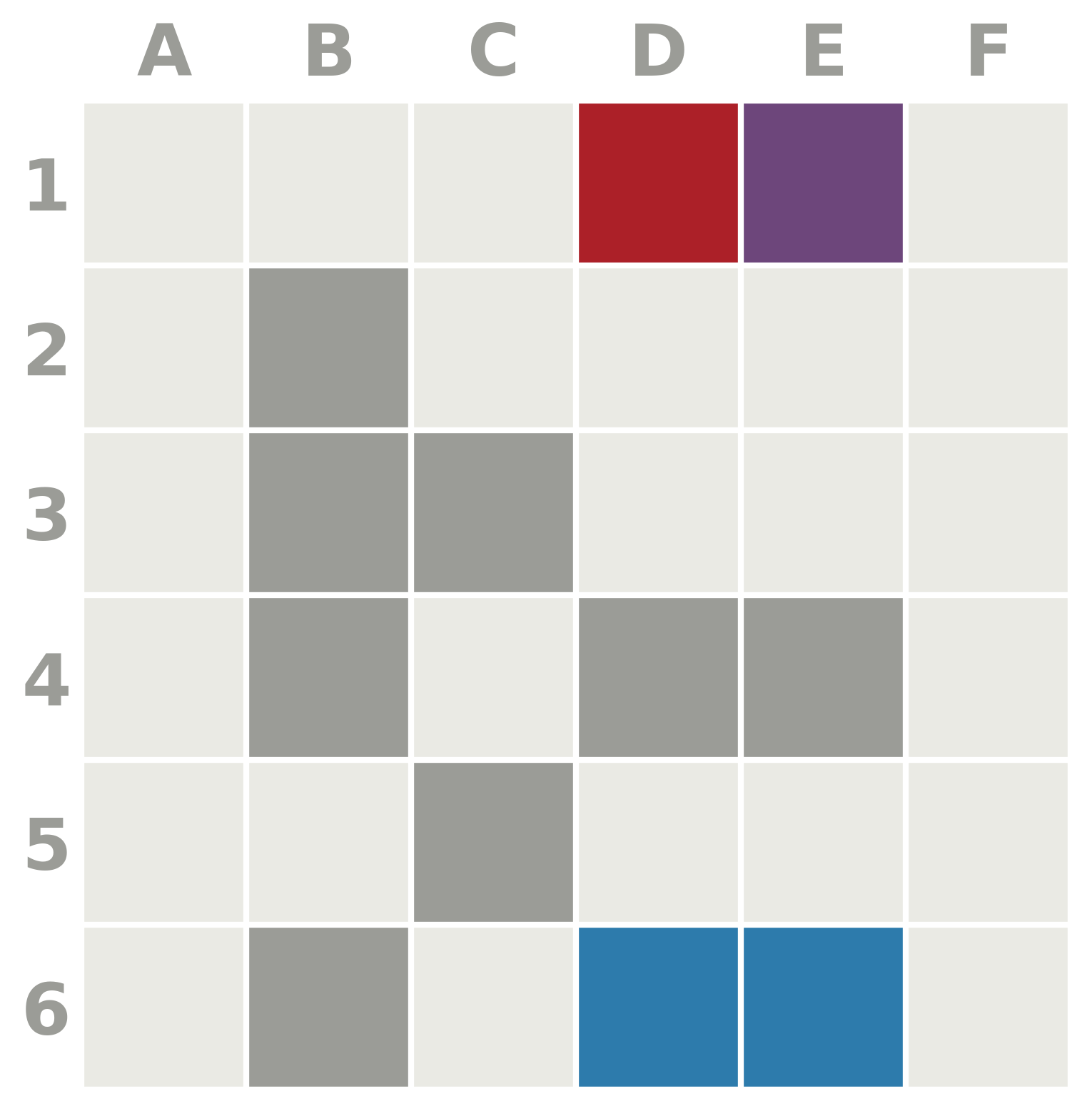}}} & \multirow[]{2}{*}{Human} & $\star$ & How many tiles in row 1 are occupied by ships? & \texttt{(++ (map (lambda x0 (++ (map (lambda y0 (== (rowL y0) 1)) (coloredTiles x0)))) (set AllColors)))} & 2.21 \\
 &  & $\epsdice{6}$ & How many tiles is the purple ship? & \texttt{(size Purple)} & 0.92 \\
\cline{2-6}
 & \multirow[]{2}{*}{CodeLlama} & $\star$ & Where is the top left part of the red ship? & \texttt{(topleft (coloredTiles Red))} & 1.92 \\
 &  & $\epsdice{6}$ & How many tiles is the red ship? & \texttt{(size Red)} & 1.52 \\
\cline{2-6}
 & \multirow[]{2}{*}{GPT-4} & $\star$ & How many tiles is the red ship? & \texttt{(size Red)} & 1.52 \\
 &  & $\epsdice{6}$ & How many tiles is the red ship? & \texttt{(size Red)} & 1.52 \\
\cline{2-6}
 & \multirow[]{2}{*}{Grammar} & $\star$ & --- & \texttt{(setSize (union (coloredTiles (color 6A)) (union (intersection (set AllTiles) (unique (coloredTiles...} & 2.98 \\
 &  & $\epsdice{6}$ & --- & \texttt{(color 1C)} & 0.97 \\
\cline{1-6}
\multirow[]{8}{*}{\shortstack[m]{Trial 18 \\ \includegraphics[width=1in]{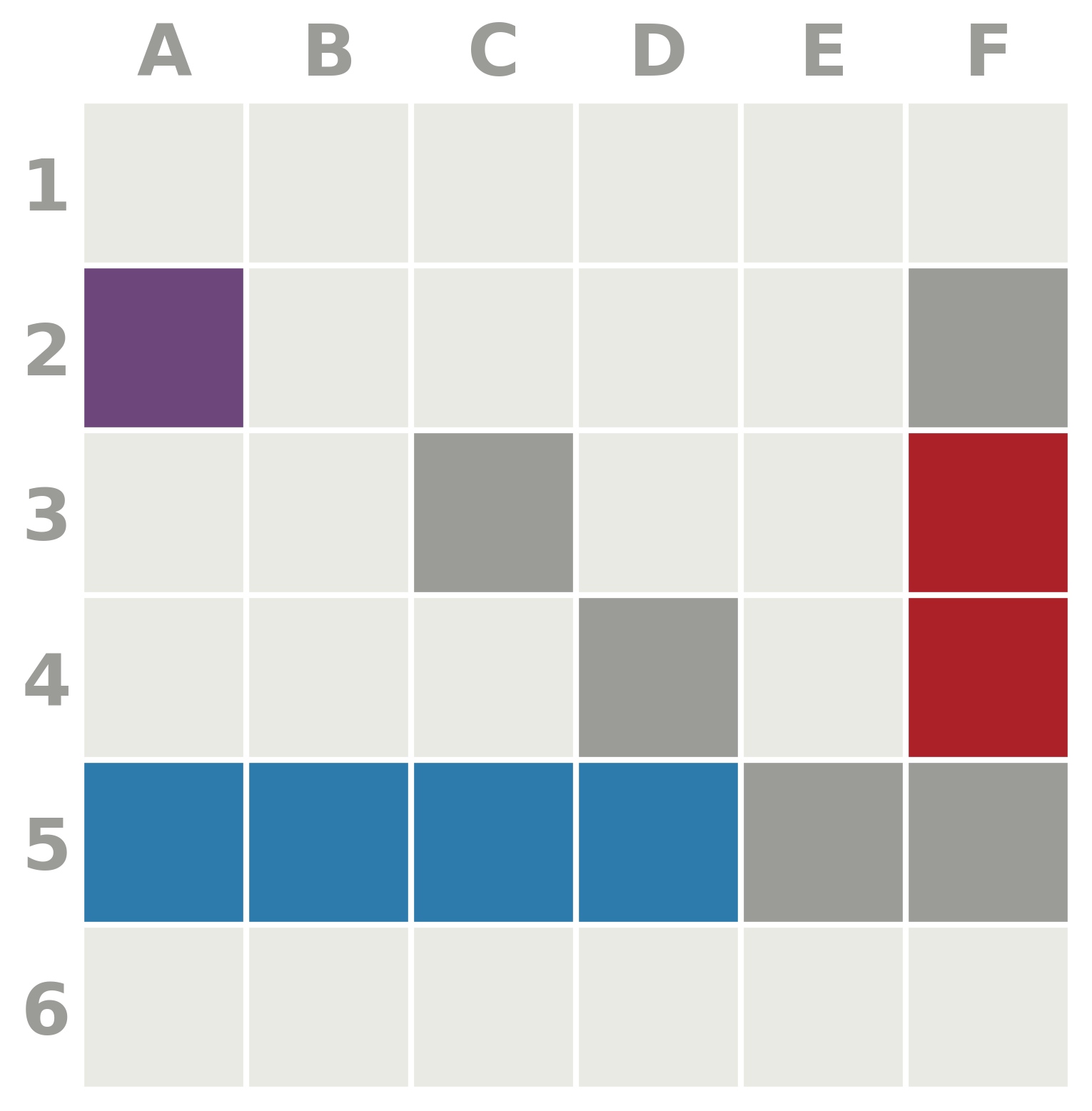}}} & \multirow[]{2}{*}{Human} & $\star$ & At what location is the bottom right part of the purple ship? & \texttt{(bottomright (coloredTiles Purple))} & 2.50 \\
 &  & $\epsdice{6}$ & How many tiles is the purple ship? & \texttt{(size Purple)} & 1.56 \\
\cline{2-6}
 & \multirow[]{2}{*}{CodeLlama} & $\star$ & How many tiles is the purple ship? & \texttt{(size Purple)} & 1.56 \\
 &  & $\epsdice{6}$ & How many tiles is the red ship? & \texttt{(size Red)} & 0.00 \\
\cline{2-6}
 & \multirow[]{2}{*}{GPT-4} & $\star$ & How many tiles is the purple ship? & \texttt{(size Purple)} & 1.56 \\
 &  & $\epsdice{6}$ & How many tiles is the purple ship? & \texttt{(size Purple)} & 1.56 \\
\cline{2-6}
 & \multirow[]{2}{*}{Grammar} & $\star$ & --- & \texttt{(setSize (coloredTiles (color 2B)))} & 2.50 \\
 &  & $\epsdice{6}$ & --- & \texttt{(size Purple)} & 1.56 \\
\cline{1-6}
\end{xltabular}

\clearpage
\newpage

\subsection{Full results}

\begin{table*}[htbp!]
\centering
\footnotesize
\begin{tabular}{llrrrrrrrrrrrr}
\toprule
 &  & \multicolumn{2}{c}{EIG} & \multicolumn{2}{c}{\% Valid} & \multicolumn{2}{c}{\% Informative} & \multicolumn{2}{c}{Program Depth} & \multicolumn{2}{c}{Program Size} & \multicolumn{2}{c}{Question Words} \\
\textbf{Model} & $k$ & $\mu$ & $\sigma_M$ & $\mu$ & $\sigma_M$ & $\mu$ & $\sigma_M$ & $\mu$ & $\sigma_M$ & $\mu$ & $\sigma_M$ & $\mu$ & $\sigma_M$ \\
\midrule
Human & 1 & 1.27 & 0.04 & 1.00 & 0.00 & 0.97 & 0.01 & 3.22 & 0.07 & 4.51 & 0.14 & 7.12 & 0.08 \\
\cline{1-14}
\multirow[t]{5}{*}{Grammar} & 1 & 0.36 & 0.00 & 1.00 & 0.00 & 0.38 & 0.00 & 3.01 & 0.00 & 5.13 & 0.01 & -- & -- \\
 & 5 & 0.98 & 0.00 & 1.00 & 0.00 & 0.89 & 0.00 & 2.74 & 0.00 & 4.07 & 0.01 & -- & -- \\
 & 10 & 1.25 & 0.00 & 1.00 & 0.00 & 0.98 & 0.00 & 2.82 & 0.00 & 4.12 & 0.01 & -- & -- \\
 & 20 & 1.49 & 0.00 & 1.00 & 0.00 & 1.00 & 0.00 & 3.08 & 0.01 & 4.62 & 0.02 & -- & -- \\
 & 50 & 1.86 & 0.00 & 1.00 & 0.00 & 1.00 & 0.00 & 3.65 & 0.01 & 5.71 & 0.04 & -- & -- \\
\cline{1-14}
\multirow[t]{5}{*}{CodeLlama-7b} & 1 & 0.65 & 0.02 & 0.75 & 0.01 & 0.45 & 0.01 & 2.64 & 0.02 & 3.24 & 0.04 & 6.66 & 0.04 \\
 & 5 & 1.24 & 0.04 & 0.99 & 0.01 & 0.90 & 0.02 & 2.49 & 0.04 & 2.89 & 0.08 & 6.77 & 0.09 \\
 & 10 & 1.55 & 0.06 & 0.99 & 0.01 & 0.97 & 0.01 & 2.36 & 0.05 & 2.58 & 0.09 & 7.10 & 0.13 \\
 & 20 & 1.83 & 0.10 & 1.00 & 0.00 & 1.00 & 0.00 & 2.34 & 0.06 & 2.46 & 0.11 & 7.61 & 0.21 \\
 & 50 & 2.31 & 0.20 & 1.00 & 0.00 & 1.00 & 0.00 & 2.58 & 0.13 & 2.69 & 0.22 & 8.56 & 0.37 \\
\cline{1-14}
\multirow[t]{5}{*}{GPT-4 (textual, few-shot)} & 1 & 0.77 & 0.02 & 0.88 & 0.01 & 0.59 & 0.01 & 2.61 & 0.02 & 3.22 & 0.04 & 6.23 & 0.03 \\
 & 5 & 1.16 & 0.04 & 0.98 & 0.01 & 0.86 & 0.02 & 2.47 & 0.05 & 2.92 & 0.09 & 6.48 & 0.05 \\
 & 10 & 1.43 & 0.05 & 1.00 & 0.00 & 0.97 & 0.01 & 2.33 & 0.06 & 2.62 & 0.12 & 6.71 & 0.07 \\
 & 20 & 1.65 & 0.09 & 1.00 & 0.00 & 1.00 & 0.00 & 2.17 & 0.04 & 2.26 & 0.06 & 6.90 & 0.11 \\
 & 50 & 2.04 & 0.19 & 1.00 & 0.00 & 1.00 & 0.00 & 2.19 & 0.07 & 2.19 & 0.07 & 7.22 & 0.14 \\
\cline{1-14}
\multirow[t]{5}{*}{GPT-4 (textual, zero-shot)} & 1 & 0.66 & 0.01 & 0.40 & 0.01 & 0.35 & 0.01 & 3.73 & 0.04 & 5.04 & 0.09 & 5.19 & 0.02 \\
 & 5 & 0.74 & 0.02 & 0.60 & 0.03 & 0.54 & 0.03 & 3.53 & 0.08 & 4.82 & 0.17 & 5.34 & 0.04 \\
 & 10 & 0.79 & 0.02 & 0.77 & 0.03 & 0.73 & 0.03 & 3.60 & 0.10 & 4.89 & 0.21 & 5.37 & 0.06 \\
 & 20 & 0.82 & 0.03 & 0.92 & 0.03 & 0.88 & 0.03 & 3.66 & 0.15 & 5.05 & 0.32 & 5.39 & 0.07 \\
 & 50 & 0.92 & 0.03 & 1.00 & 0.00 & 1.00 & 0.00 & 3.42 & 0.11 & 4.33 & 0.14 & 5.36 & 0.11 \\
\cline{1-14}
\multirow[t]{5}{*}{GPT-4 (grid, few-shot)} & 1 & 0.62 & 0.02 & 0.85 & 0.01 & 0.49 & 0.01 & 2.72 & 0.02 & 3.42 & 0.04 & 6.06 & 0.03 \\
 & 5 & 1.00 & 0.03 & 0.96 & 0.01 & 0.77 & 0.02 & 2.56 & 0.05 & 3.10 & 0.10 & 6.31 & 0.06 \\
 & 10 & 1.18 & 0.05 & 0.99 & 0.01 & 0.87 & 0.03 & 2.41 & 0.06 & 2.80 & 0.13 & 6.58 & 0.08 \\
 & 20 & 1.38 & 0.07 & 1.00 & 0.00 & 0.92 & 0.03 & 2.27 & 0.08 & 2.50 & 0.16 & 6.84 & 0.12 \\
 & 50 & 1.64 & 0.14 & 1.00 & 0.00 & 1.00 & 0.00 & 2.25 & 0.07 & 2.39 & 0.12 & 7.22 & 0.24 \\
\cline{1-14}
\multirow[t]{5}{*}{GPT-4 (grid, zero-shot)} & 1 & 0.56 & 0.01 & 0.55 & 0.01 & 0.39 & 0.01 & 3.30 & 0.03 & 4.49 & 0.06 & 5.85 & 0.04 \\
 & 5 & 0.79 & 0.02 & 0.88 & 0.02 & 0.80 & 0.02 & 3.24 & 0.05 & 4.42 & 0.10 & 5.71 & 0.07 \\
 & 10 & 0.89 & 0.02 & 0.96 & 0.01 & 0.93 & 0.02 & 3.20 & 0.06 & 4.33 & 0.12 & 5.82 & 0.09 \\
 & 20 & 0.94 & 0.01 & 1.00 & 0.00 & 1.00 & 0.00 & 3.16 & 0.08 & 4.26 & 0.16 & 5.86 & 0.11 \\
 & 50 & 0.99 & 0.02 & 1.00 & 0.00 & 1.00 & 0.00 & 3.19 & 0.15 & 4.36 & 0.30 & 5.83 & 0.21 \\
\cline{1-14}
\multirow[t]{5}{*}{GPT-4 (visual, few-shot)} & 1 & 0.54 & 0.01 & 0.80 & 0.01 & 0.46 & 0.01 & 3.02 & 0.02 & 4.01 & 0.04 & 5.69 & 0.03 \\
 & 5 & 0.89 & 0.03 & 0.91 & 0.01 & 0.75 & 0.02 & 2.97 & 0.06 & 3.92 & 0.12 & 5.92 & 0.07 \\
 & 10 & 1.08 & 0.04 & 0.99 & 0.01 & 0.92 & 0.02 & 2.81 & 0.07 & 3.60 & 0.16 & 6.16 & 0.10 \\
 & 20 & 1.29 & 0.06 & 1.00 & 0.00 & 0.98 & 0.02 & 2.56 & 0.09 & 3.07 & 0.19 & 6.56 & 0.15 \\
 & 50 & 1.56 & 0.12 & 1.00 & 0.00 & 1.00 & 0.00 & 2.42 & 0.13 & 2.72 & 0.25 & 7.03 & 0.24 \\
\cline{1-14}
\multirow[t]{5}{*}{GPT-4 (visual, zero-shot)} & 1 & 0.34 & 0.01 & 0.58 & 0.01 & 0.25 & 0.01 & 2.18 & 0.02 & 2.28 & 0.03 & 1.11 & 0.01 \\
 & 5 & 0.73 & 0.03 & 0.70 & 0.02 & 0.56 & 0.03 & 2.18 & 0.04 & 2.30 & 0.07 & 1.03 & 0.01 \\
 & 10 & 0.88 & 0.04 & 0.80 & 0.03 & 0.71 & 0.03 & 2.07 & 0.03 & 2.10 & 0.04 & 1.00 & 0.00 \\
 & 20 & 0.99 & 0.05 & 1.00 & 0.00 & 0.92 & 0.03 & 2.07 & 0.04 & 2.09 & 0.05 & 1.00 & 0.00 \\
 & 50 & 1.19 & 0.07 & 1.00 & 0.00 & 1.00 & 0.00 & 2.22 & 0.11 & 2.31 & 0.15 & 1.00 & 0.00 \\
\cline{1-14}
\multirow[t]{5}{*}{GPT-4 (no board, few-shot)} & 1 & 0.60 & 0.02 & 0.68 & 0.01 & 0.43 & 0.01 & 3.08 & 0.03 & 4.12 & 0.07 & 6.28 & 0.03 \\
 & 5 & 0.98 & 0.03 & 0.98 & 0.01 & 0.89 & 0.02 & 3.01 & 0.07 & 3.97 & 0.13 & 6.24 & 0.08 \\
 & 10 & 1.19 & 0.05 & 1.00 & 0.00 & 0.97 & 0.01 & 2.82 & 0.07 & 3.59 & 0.15 & 6.31 & 0.12 \\
 & 20 & 1.38 & 0.08 & 1.00 & 0.00 & 0.98 & 0.02 & 2.73 & 0.11 & 3.42 & 0.24 & 6.80 & 0.19 \\
 & 50 & 1.75 & 0.18 & 1.00 & 0.00 & 1.00 & 0.00 & 2.72 & 0.19 & 3.28 & 0.40 & 7.64 & 0.36 \\
\cline{1-14}
\multirow[t]{5}{*}{GPT-4 (no board, zero-shot)} & 1 & 0.65 & 0.01 & 0.69 & 0.01 & 0.50 & 0.01 & 3.37 & 0.03 & 4.67 & 0.05 & 6.55 & 0.02 \\
 & 5 & 0.81 & 0.02 & 0.94 & 0.01 & 0.81 & 0.02 & 3.32 & 0.05 & 4.59 & 0.10 & 6.27 & 0.04 \\
 & 10 & 0.88 & 0.02 & 1.00 & 0.00 & 0.92 & 0.02 & 3.33 & 0.07 & 4.61 & 0.14 & 6.26 & 0.06 \\
 & 20 & 0.93 & 0.03 & 1.00 & 0.00 & 0.94 & 0.02 & 3.47 & 0.12 & 4.89 & 0.24 & 6.39 & 0.10 \\
 & 50 & 1.01 & 0.02 & 1.00 & 0.00 & 1.00 & 0.00 & 3.89 & 0.24 & 5.72 & 0.49 & 6.83 & 0.19 \\
\bottomrule
\end{tabular}

\caption{Full statistics for all models and values of $k$. $\mu$ and $\sigma_M$ denote sample mean and standard error, respectively, and are computed across all board contexts. Questions that translated to a parseable program are considered Valid, and those that achieved $\EIG > 0$ are considered Informative. Program Depth and Size refer to the depth and number of nodes of the program abstract syntax tree. Question Words measures the number of words in the natural language question.}
\label{apx:tab:full_results}
\end{table*}

\clearpage
\newpage
\section{Prompts}

Our model procedurally constructs few-shot LLM prompts to elicit task-relevant questions and translations. There are two prompt formats: one for question-generation and one for translation. Each prompt is structured as a series of messages conveying instructions, few-shot examples, or information about the target task. In some cases, the format of the message varies depending on the modality of the board representation (textual, grid, or visual).

Following emerging conventions around APIs for conversational AI models, each component is labeled with a \textit{role}. \texttt{System} provides general high-level instructions; \texttt{User} indicates inputs from a user; and \texttt{Assistant} indicates responses generated from the model. These role labels are either passed as metadata (for GPT-4) or prepended to the text of each message (CodeLlama). Note that the purpose of these role labels is to mock illustrate a desired interaction pattern; the LLM only generates text at the end of the conversation.

\subsection{Question generation prompt}

\paragraph{Instructions} The prompt begins with a system message explaining the role of the LLM (``You are a game-playing agent...''). This is followed by a set of general instructions describing the Battleship task. Finally, one of three modality-specific messages is given to describe the format of the board representation.

\paragraph{Few-shot examples} Next, to illustrate the desired behavior, we provide several few-shot examples of boards and questions. Concretely, we randomly choose 3 boards that are not the target board, and randomly choose 10 questions for each board from the human data. (All sampling is done without replacement.) In the ``no board'' condition, the board representation is omitted, but the example questions are still present. In the zero-shot condition, the entire block beginning with ``Here are some examples...'' is omitted.

\paragraph{Target board} Finally, the prompt concludes with the target board in order to elicit a new question from the LLM. In the ``no board'' condition, the transition message (``Now, it's your turn...'') and the target board are both omitted, so that the prompt effectively reduces to a list of example questions that the LLM extends without any knowledge of the board.

\begin{systembox}
You are a game-playing agent. Read the game instructions and examples carefully. Respond with a single question that can be answered with one word. Do not include any other explanation or prose.
\end{systembox}

\begin{userbox}
You are playing the board game Battleship. There are three ships on the board: Red, Blue, and Purple. Ships are oriented either horizontally or vertically and can be 2, 3, or 4 tiles in length. The board is a 6x6 grid, with numbered rows 1, 2, 3, 4, 5, 6 and lettered columns A, B, C, D, E, F. Coordinates are specified as a row, column pair. For example, 2-C is the tile in row 2, column C.\\

You will be given a partially-revealed game board. Your task is to ask a single question that will help you gain information about the position of the remaining hidden ships on the board. You can ask any question, but it must be answerable with a single word answer.
\end{userbox}

\begin{tcbraster}[raster columns=3, raster equal height, raster column skip=5mm]
    \begin{userbox}[User (textual)]
        The board is represented as a textual description.
    \end{userbox}
    \begin{userbox}[User (grid)]
        The board is represented as a grid with the following symbols:\\
        
        H: Hidden\\
        W: Water\\
        R: Red ship\\
        B: Blue ship\\
        P: Purple ship\\
    \end{userbox}
    \begin{userbox}[User (visual)]
        The board is represented as an image, with light gray indicating hidden tiles, dark gray indicating water tiles, and red, blue and purple indicating ship tiles.
    \end{userbox}
\end{tcbraster}

\begin{userbox}
Here are some examples of questions from other agents about different boards.
\end{userbox}

\begin{tcolorbox}[enhanced, 
                  frame hidden,
                  colback=gray!20!white,
                  overlay={
                    \draw[decoration={brace,amplitude=10pt},decorate,thick] 
                      ([xshift=-5pt]frame.south west) -- 
                      ([xshift=-5pt]frame.north west) 
                      node[midway,xshift=-10pt,anchor=east] {\texttt{3x}};
                  }]
    \begin{tcbraster}[raster columns=3, raster equal height, raster column skip=5mm]
        \begin{userbox}[User (textual)]

            \vspace{1ex}
            2-C is a water tile.\\
            2-E is a water tile.\\
            3-C is a purple ship tile.\\
            4-D is a water tile.\\
            5-B is a water tile.\\
            6-E is a water tile.
        \end{userbox}
        \begin{userbox}[User (grid)]

            \vspace{1ex}
            \hspace{0.5em} A B C D E F\\
            1 H H H H H H\\
            2 H H W H W H\\
            3 H H P H H H\\
            4 H H H W H H\\
            5 H W H H H H\\
            6 H H H H W H
        \end{userbox}
        \begin{userbox}[User (visual)]

            \vspace{2ex}
            \includegraphics[width=1in]{figures/boards/board_2.png}
        \end{userbox}
    \end{tcbraster}
    \begin{tcolorbox}[enhanced, 
                      frame hidden,
                      colback=gray!20!white,
                      left=4em,
                      top=0em,
                      bottom=0em,
                      right=0em,
                      overlay={
                        \draw[decoration={brace,amplitude=10pt},decorate,thick] 
                          ([xshift=40pt]frame.south west) -- 
                          ([xshift=40pt]frame.north west) 
                          node[midway,xshift=-10pt,anchor=east] {\texttt{N=10}};
                      }]
        \begin{assistantbox}
          At what location is the top left part of the red ship?
        \end{assistantbox}
        \begin{assistantbox}
          Is the red ship horizontal?
        \end{assistantbox}
        \begin{assistantbox}
          Is there any ship in column F?
        \end{assistantbox}
    \end{tcolorbox}
\end{tcolorbox}

\begin{userbox}
Now, it's your turn. Here is your board:
\end{userbox}

\begin{tcbraster}[raster columns=3, raster equal height, raster column skip=5mm]
    \begin{userbox}[User (textual)]

        \vspace{1ex}
        1-B is a purple ship tile.\\
        1-C is a water tile.\\
        1-E is a water tile.\\
        2-D is a red ship tile.\\
        2-E is a blue ship tile.\\
        3-B is a water tile.\\
        ...
    \end{userbox}
    \begin{userbox}[User (grid)]

        \vspace{1ex}
        \hspace{0.5em} A B C D E F\\
        1 H P W H W H\\
        2 H H H R B H\\
        3 H W H H H H\\
        4 W H W H H W\\
        5 H H W W H H\\
        6 H W H H H H
    \end{userbox}
    \begin{userbox}[User (visual)]

        \vspace{2ex}
        \includegraphics[width=1in]{figures/boards/board_16.png}
    \end{userbox}
\end{tcbraster}

\begin{assistantbox}
...
\end{assistantbox}

\clearpage
\newpage
\subsection{Translation prompt}

\paragraph{Instructions} The prompt also begins with a system message containing a general instructions describing the Battleship task and explaining the assistant's role as as translator. The Battleship task instructions are identical to the ones given in the question-generation prompt.

\paragraph{Few-shot examples} The body of the prompt consists of 12 pairs of (language, code) examples illustrating the desired translation behavior. These examples are randomly sampled without replacement from the human data. We exclude any examples pertaining to the same board as the one targeted during question-generation.

\paragraph{Target language} Finally, the prompt concludes with a target question in language, which the LLM translates into code.

\begin{systembox}
You are playing the board game Battleship. There are three ships on the board: Red, Blue, and Purple. Ships are oriented either horizontally or vertically and can be 2, 3, or 4 tiles in length. The board is a 6x6 grid, with numbered rows 1, 2, 3, 4, 5, 6 and lettered columns A, B, C, D, E, F. Coordinates are specified as a row, column pair. For example, 2-C is the tile in row 2, column C.\\

Your task is to translate each of the user's questions into a query program.
\end{systembox}

\begin{tcolorbox}[enhanced, 
                  frame hidden,
                  colback=gray!20!white,
                  overlay={
                    \draw[decoration={brace,amplitude=10pt},decorate,thick] 
                      ([xshift=-5pt]frame.south west) -- 
                      ([xshift=-5pt]frame.north west) 
                      node[midway,xshift=-10pt,anchor=east] {\texttt{N=12}};
                  }]
    \begin{userbox}
    How many tiles is the red ship?
    \end{userbox}
    \begin{assistantbox}
    (size Red)
    \end{assistantbox}
    \begin{userbox}
    Do the red ship and the purple ship touch?
    \end{userbox}
    \begin{assistantbox}
    (touch Red Purple)
    \end{assistantbox}
    \begin{userbox}
    Is there a ship at 1F?
    \end{userbox}
    \begin{assistantbox}
    (not (== (color 1F) Water))
    \end{assistantbox}
    \begin{userbox}
    Is the blue ship horizontal?
    \end{userbox}
    \begin{assistantbox}
    (== (orient Blue) H)
    \end{assistantbox}
    \begin{userbox}
    How many ships are horizontal?
    \end{userbox}
    \begin{assistantbox}
    (++ (map (lambda x0 (== (orient x0) H)) (set AllColors)))
    \end{assistantbox}
\end{tcolorbox}
\begin{userbox}
Are there more horizontal ships than vertical ships?
\end{userbox}
\begin{assistantbox}
...
\end{assistantbox}

\fi
\end{document}